\documentclass{article} % For LaTeX2e
\usepackage{iclr2023_conference,times}

\usepackage{amsmath,amsfonts,bm}

\def\eqref#1{equation~\ref{#1}}
\def\1{\bm{1}}

\DeclareMathAlphabet{\mathsfit}{\encodingdefault}{\sfdefault}{m}{sl}
\SetMathAlphabet{\mathsfit}{bold}{\encodingdefault}{\sfdefault}{bx}{n}

\usepackage{marvosym}
\usepackage[utf8]{inputenc} % allow utf-8 input
\usepackage[T1]{fontenc}    % use 8-bit T1 fonts
\usepackage{hyperref}       % hyperlinks
\usepackage{url}            % simple URL typesetting
\usepackage{booktabs}       % professional-quality tables
\usepackage{amsfonts}       % blackboard math symbols
\usepackage{nicefrac}       % compact symbols for 1/2, etc.
\usepackage{microtype}      % microtypography
\usepackage{xcolor}         % colors
\usepackage{graphicx}
\usepackage{float}
\usepackage{caption}
\usepackage{subcaption}
\usepackage{verbatim}
\usepackage{wrapfig}
\usepackage{amsmath}
\usepackage{paralist}

\usepackage[justification=centering]{caption}
\mathchardef\mhyphen="2D
\title{Causal Information Bottleneck Boosts Adversarial Robustness of Deep Neural Network}
\author{Huan Hua,
        Jun Yan \Letter,
        Xi Fang,
        Weiquan Huang,
        Huilin Yin, 
        Wancheng Ge % <-this % stops a space
\thanks{Huan Hua, Jun Yan, Weiquan Huang, Huilin Yin, and Wancheng Ge are with the SEIEE college of Tongji University, Shanghai, China. Xi Fang is with the SmartMore. Co. Ltd.}% <-this % stops a space
\thanks{Jun Yan (yanjun@tongji.edu.cn) is the corresponding author.}}
\begin{document}
\maketitle

\begin{abstract}
The information bottleneck (IB) method is a feasible defense solution against adversarial attacks in deep learning. However, this method suffers from the spurious correlation, which leads to the limitation of its further improvement of adversarial robustness. In this paper, we incorporate the causal inference into the IB framework to alleviate such a problem. Specifically, we divide the features obtained by the IB method into robust features (content information) and non-robust features (style information) via the instrumental variables to estimate the causal effects. With the utilization of such a framework, the influence of non-robust features could be mitigated to strengthen the adversarial robustness. We make an analysis of the effectiveness of our proposed method. The extensive experiments in MNIST, FashionMNIST, and CIFAR-10 show that our method exhibits the considerable robustness against multiple adversarial attacks. Our code would be released.
\end{abstract}

\section{Introduction}

With the continuous improvement of computing power and data availability, the deep neural networks (DNNs) have made breakthroughs in many fields, such as image classification~\cite{imageClassification}, object detection~\cite{objectDetection}, machine translation~\cite{machineTranslation}, natural language understanding~\cite{BERT}, and so on. In DNNs, feature maps in the middle layers are treated as compression code $Z$. However, many studies in recent years have shown that DNNs are susceptible to adversarial examples~\cite{existOfAdverSzegedy, existOfAdverGoodfellow, TowardsAdver}. In the field of computer vision, adversarial examples which manipulate a small number of image pixels without the change of semantic representation can deceive DNNs to make false predictions~\cite{reviewOfAdverXu}. It would be a huge threat to autonomous driving~\cite{driving}, face recognition~\cite{FaceRec}, and daily shopping~\cite{onlineShopping}. Therefore, the security of deep neural networks has become a significant concern.

The phenomenon of the adversarial vulnerability can be regarded as the overfitting problem of DNNs~\cite{explaingingOfAdverGoodfellow}. Moreover, the defect of the decision boundaries in DNNs leads to the possibility of adversarial attacks~\cite{explaingingOfAdverGoodfellow2}. As an effective regularization method, the information bottleneck (IB) theory can help reduce the adversarial empirical risk and approximate a better decision boundary. The IB theory is an extension of Shannon's rate-distortion theory~\cite{IBTishby}. Its goal is to find an optimal compression code for the target random variable, which maximizes the mutual information between the target random variable and the compression code, and minimizes the mutual information between the source random variable and the compression code. The IB method is believed to be useful to explain the operation and principle of DNNs, and the mechanism of the information compression in the method of IB helps the DNNs extract the representative features~\cite{IBTishby, OpeningBBTishby, DLTishby}. Many pieces of follow-up work have been inspired, including the exploration of the relationship between adversarial robustness and  IB theory~\cite{InformationDropout,VIB,CEB2}.  In recent years, many theoretical analyses have been proposed, and many models and algorithms have also been shown to be effective, including Information Dropout~\cite{InformationDropout} and Variation IB (VIB)~\cite{VIB}, Diesentangled IB~\cite{DisenIB}, and so on~\cite{DistillingRobustandNonRubust,CEB2,variationalGlass}.

Although some IB-based methods improve the adversarial robustness of the model to a certain extent, there is a key problem with these methods: the existing IB methods~\cite{OpeningBBTishby, DLTishby, VIB, CEB} do not pay attention to the problem of the spurious correlation that the robust features and the non-robust features would be entangled with each other. The learning bias on the fragile and incomprehensible features (non-robust features) would lead to the adversarial vulnerability. In contrast, mankind with the ability of causal inference~\cite{GenericVisualPerception,learningCausalMechanisms} can normally recognize the visual adversarial examples correctly by identifying and peeling those unrelated factors~\cite{CausalityPearl}. With the utilization of causal inference, DNN models would focus more on robust features~\cite{CiiV}.
\begin{figure}[!t]
\centering
\includegraphics[width=5.0in,height=1.5in]{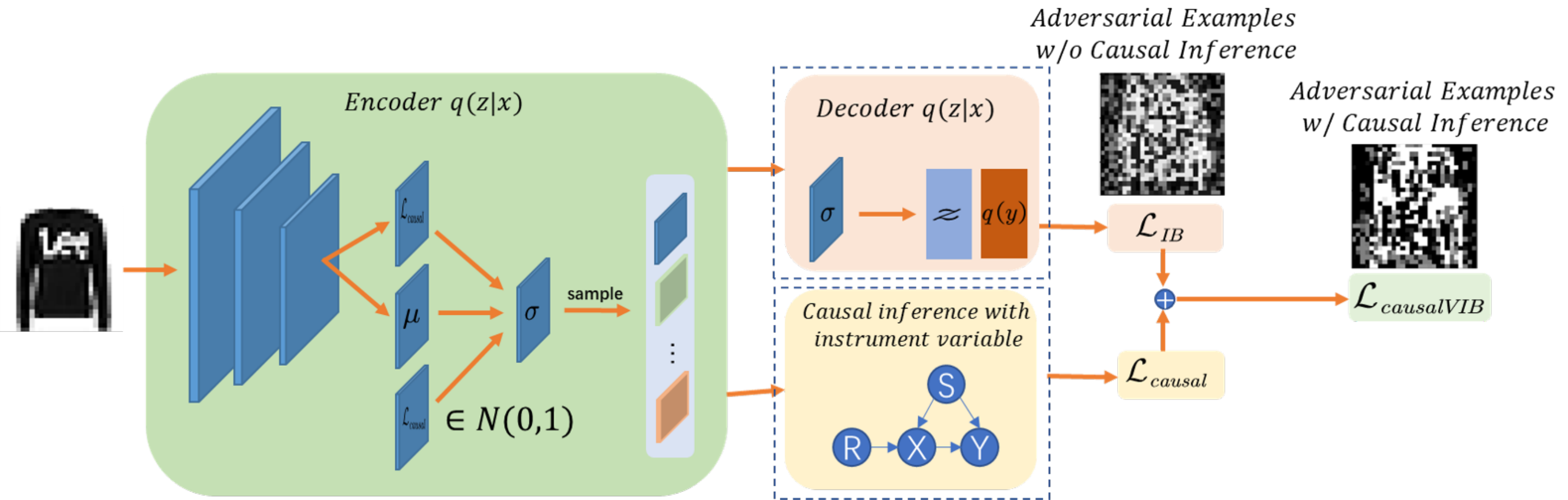} 
%\flushleft
\captionsetup{justification=raggedright}
\caption{The proposed CausalIB framework is divided into two modules, IB and causal inference. The IB module is used for overall regularization, and the causal inference module is used to distinguish robust features from non-robust features. With the utilization of causal inference, the proposed CausalIB can learn more structured information than models using only IB module.}
\label{fig:image1}
\end{figure}
To alleviate the problem of the entanglement between the robust features and the non-robust features in the IB framework, we propose a causal intervention method. In this study, we assume that the style information of an image is a non-robust feature, while the content information is a robust feature. Since the style information cannot be observed, we introduce instrumental variables to help remove the influence of style information. In the experiment, we validate the effectiveness of the proposed causal inference IB method under various adversarial attacks such as Fast Gradient Sign Method (FGSM)~\cite{explaingingOfAdverGoodfellow}, PGD~\cite{TowardsAdver} on the MNIST~\cite{MNISTref}, FashionMNIST~\cite{fashionMNISTref} and CIFAR-10~\cite{krizhevsky2009learning} datasets. The empirical results validate the effectiveness of our method to boost the adversarial robustness.

Our contributions are as follows:

1. We utilize the causal theory to analyze the cause of the adversarial vulnerability of deep learning models and the feasibility of improving robustness through instrumental variables;

2. We introduce a model called CausalIB, which uses the IB method with the assistance of causality to extract features and causal inference to disentangle the robust features and non-robust features.

3. The extensive experiments on various settings of MNIST, FashionMNIST, and CIFAR-10 show that CausalIB is robust against adversarial attacks.

The remainder of this paper is organized as follows.
First, we make a literature review in Section 2. In Section 3, we introduce the IB theory and analyze the reasons for the existence of adversarial examples from the perspective of causal theory. Moreover, the approach is illustrated in Section 3. The experiment's detail and result will be described in Section 4. Finally, the conclusion of the research is given in Section 5.
    
\section{Related Work}
In this section, we briefly review the current state of research on adversarial robustness, prior works on IB methods, and related causal inference methods.

\subsection{Adversarial Robustness}
The existing methods for dealing with adversarial examples mainly include three directions~\cite{ThratOfAdverAttack}: preprocessing methods for the defense~\cite{miyato2016adversarial,zheng2016improving,shin2017jpeg}, improvement on the neural network structures~\cite{rifai2011contractive,bai2017alleviating,hinton2015distilling}, and the utilization of external models when classifying  unseen examples~\cite{akhtar2018defense,lee2017generative}. Currently, the most effective strategy is adversarial training~\cite{explaingingOfAdverGoodfellow,EnsembleAdversarialTraining}. Adversarial training can be regarded as a method of data augmentation, its adversarial robustness largely depends on the coverage of adversarial examples during the training process. However, the study of Moosavi Dezhouni et al.~\cite{UniversalAdverPertu} found that even a well-trained defense network can still obtain other effective adversarial examples through computation which brings new difficulties for the adversarial defense. The methods without the utilization of adversarial training such as preprocessing methods~\cite{FeatureDenoising, ImproveWithDenoising, KeepBadGuysOut}, data randomization methods~\cite{TheoreticalEvidenceAdver, RS, AdverSemanticSeg}, or IB methods~\cite{IBTishby, OpeningBBTishby} also provide insight for adversarial robustness. Among them, Das et al.~\cite{KeepBadGuysOut} utilized the method of compression to remove the high-frequency components from images to improve robustness. Some pieces of seminal work~\cite{AdverSemanticSeg, RandomAdver} show that data randomization has a role in reducing the fooling rates of the networks. Initially, the IB theory was seen as an interpretive work for DNNs~\cite{IBTishby}. However, it has also been found that the IB method is an effective regularization method due to its trade-off between prediction performance and model compression, which can be used to improve the adversarial robustness~\cite{VIB,CEB2,InformationDropout,DisenIB}.
\subsection{IB Method}
In recent years, IB methods have been widely used to improve model robustness. Achille et al.~\cite{InformationDropout} improved the Dropout method with the utilization of the IB framework to reduce the sensitivity of the model to perturbations. A similar method is delivered by Kim et al.~\cite{DistillingRobustandNonRubust} that the robust and the non-robust neurons could be separated according to the different encoding values reacting to the noises. Different from the above methods, Alemi et al.~\cite{VIB} proposed a variational approximation method to optimize the IB, using reparameterization trick~\cite{rePara} for efficient training and demonstrating the effectiveness against adversarial examples. Fischer et al.~\cite{CEB2} postulated that the vulnerability of neural networks stems from the fact that the model retains too much information about the training data, resulting in weakness under the adversarial attacks. Considering that the essence of the IB is the restriction of the complexity of representation learning, the proposal of the conditional entropy-based IB (CEB) strengthened the adversarial robustness and generalization ability of the IB framework~\cite{CEB}. \textcolor{black}{
A recent study~\cite{closerlook} postulates that the information bottleneck methods based on the variational inference would suffer from gradient obfuscation due to the non-smooth loss surfaces durint the optimization process. Such a challenge inspires us to explore the feasible method to promote adversarial robustness of IB methods.} Besides, DiesenIB~\cite{DisenIB} implements the IB method from the perspective of supervised disentanglement. In the proposed DiesenIB, the information is not compressed, no compression means no loss of prediction performance, and disentanglement helps the model to achieve a better performance in out-of-distribution (OOD) and adversarial defense. Through improving VIB, Sinha~\cite{DIBS} obtained the diversity of output prediction, which is required for multimodal data modeling, and achieved certain results in sparse training data and uncertainty estimation for OOD detection, etc.
\subsection{Causal Inference}
Human perception is robust to adversarial perturbations due to the ability of causal inference~\cite{CausalViewRobust, CausalityPearl,CausalInference}. Many recent works have shown in various aspects that causal models are suitable methods for parametric reasoning in complex systems~\cite{counterfactualFairness}, and can even be utilized to interpret the deep learning models~\cite{TowardCausalRepresentation}. Recently, Zhang et al.~\cite{AdverThrougnLen} proposed a method called the adversarial distribution alignment, which attempts to explain the existence of adversarial examples from the perspective of causality. Such a method removes spurious correlations by eliminating the difference between natural and adversarial distributions. A similar approach called causal manipulation augmented model~\cite{CausalViewRobust} aims to improve the robustness of DNNs to unseen adversarial perturbations by explicitly modeling the perturbations from a causal view. The stable learning methods are delivered~\cite{CausalInferencebyUsingInvariantPrediction, StablePrediction} to reduce the accuracy variance of the model under various sample distributions via the causal methods. Most relevant to our work is the causal intervention by instrumental variable (CiiV) model~\cite{CiiV}. \textbf{However, the CiiV model uses retinotopic sampling to intervene in the image layer (original image layer). The difference between our proposed method and the previous work~\cite{CiiV} is that we use additive noise as the instrumental variable in the concept layer (intermediate feature layer) rather than in the input layer.}
\section{Approach}
\label{sec:blind}
\subsection{Information Bottleneck}
The core idea of the IB theory is information compression, that is, to maximize the mutual information between the target random variable and the compression code, and minimize the mutual information between the source random variable and the compression code. The loss function of IB is defined as Eq.\ref{originIBLoss}:
\begin{equation}\label{originIBLoss}
\mathcal{L} _{IB}=-I\left( Z;Y \right) +\alpha I\left( Z;X\right),
\end{equation}
where $I(Z;Y)$ is the mutual information between the target random variable $Y$ and the compression code $Z$, and $I(Z;X)$ is the mutual information between the source random variable $X$ and $Z$. The IB restricts the correlation between $X$ and $Z$, which is an effective regularization method to reduce the empirical risk. In DNNs, feature maps in the middle layers are treated as compression codes. However, it is not easy to calculate mutual information in DNNs. In this regard, the variational inference is added into the IB framework to construct the variational IB (VIB) model~\cite{VIB}. The lower bound of $I(Z,Y)$ and the upper bound of $I(Z,X)$ are formalized as Eq.\ref{LowerBound} and Eq.\ref{UpperBound}:
\begin{equation}\label{LowerBound}
I\left( Z,Y \right) \,\,\ge \,\,\int{dxdydz\,\,p\left( x \right) p\left( y|x \right) p\left( z|x \right) \log q\left( y|z \right)}
=\mathbb{E} _{q\left( y,t \right)}\log p\left( y|t \right),
\end{equation}
\begin{equation}\label{UpperBound}
I\left( Z,X \right) \,\,\le \,\,\int{dxdz\,\,p\left( x \right) p\left( z|x \right) \log \frac{p\left( z|x \right)}{r\left( z \right)}},
\end{equation}
where $r(z)$ is the prior probability of the latent variable $Z$, which is set to a multi-dimensional Gaussian distribution with mean 0 and variance 1.
% Thus, the IB loss function can be formalized as Eq.\ref{IBLoss}:
% \begin{equation}
%     \begin{aligned} I(Z, Y)-\alpha I(Z, X) \geq & \int d x d y d z p(x) p(y \mid x) p(z \mid x) \log q(y \mid z) \\ &-\alpha \int d x d z p(x) p(z \mid x) \log \frac{p(z \mid x)}{r(z)}=\mathcal{L} _{IB}. \end{aligned}\label{IBLoss}
% \end{equation}
Reparameterization trick~\cite{rePara} is used to estimate the mutual information. Unlike deterministic models, with the reparameterization trick, VIB learns not only deterministic features in feature layers but also a whole multi-dimensional Gaussiann distribution.

The mean of samples of the distribution would be taken by VIB, and the loss function can be written as Eq.\ref{newIBLoss}:
\begin{equation}\label{newIBLoss}
    \begin{aligned}
        \mathcal{L} _{IB}   &\approx \frac{1}{N}\sum_{n=1}^N{\left[ \int{d}zp\left( z\mid x_n \right) \log q\left( y_n\mid z \right) -\alpha p\left( z\mid x_n \right) \log \frac{p\left( z\mid x_n \right)}{r(z)} \right]} \\
                            &=\frac{1}{N}\sum_{n=1}^N{\mathbb{E} _{\epsilon \sim p(\epsilon )}}\left[ -\log q\left( y_n\mid f\left( x_n,\epsilon \right) \right) \right] +\alpha \mathrm{KL}\left[ p\left( Z\mid x_n \right) ,r(Z) \right] ,
    \end{aligned}
\end{equation}
%     \mathcal{L} _{IB}\approx \frac{1}{N}\sum_{n=1}^N{\left[ \int{d}zp\left( z\mid x_n \right) \log q\left( y_n\mid z \right) -\alpha p\left( z\mid x_n \right) \log \frac{p\left( z\mid x_n \right)}{r(z)} \right]}
% \\
% =\frac{1}{N}\sum_{n=1}^N{\mathbb{E} _{\epsilon \sim p(\epsilon )}}\left[ -\log q\left( y_n\mid f\left( x_n,\epsilon \right) \right) \right] +\alpha \mathrm{KL}\left[ p\left( Z\mid x_n \right) ,r(Z) \right] ,
% \end{equation}
% In practice, the above objective is approximated as Eq.\ref{practicalVIB}:
% \begin{equation}\label{practicalVIB}
% \mathcal{L}_{I B}=\frac{1}{N} \sum_{n=1}^{N} \mathbb{E}_{\epsilon \sim p(\epsilon)}\left[-\log q\left(y_{n} \mid f\left(x_{n}, \epsilon\right)\right)\right]+\alpha \mathrm{KL}\left[p\left(Z \mid x_{n}\right), r(Z)\right],
% \end{equation}
where the former term is the cross entropy, and the latter term is the Kullback–Leibler (KL) divergence.
\subsection{A Causal View of Existence of Adversarial Examples}
The existing neural networks would learn a spurious correlation between data and predictions which induces their vulnerabilities~\cite{AdverThrougnLen,CiiV,StablePrediction}. We postulate that these promiscuous, spuriously correlated features, such as background features, color features, or even the bias of camera angles may lead to the adversarial vulnerability. Humans are not affected by such disturbances due to the ability of causal inference including observation, intervention, and counterfactual~\cite{CausalityPearl, ElementsofCausalInference}. Therefore, we turn to causal theory to mitigate the spurious correlations that existed in the learning process of DNNs.

Taking a figure consisting of a dog and a grass background as an example, humans do not interfere with the information of grass when recognizing the dog, but DNNs will learn both the dog information and the grass information. In the process of feature learning, DNN models simply observe all the data without removing the spurious correlations between the features~\cite{AdverThrougnLen}, 
and the model is quite sensitive to these features with spurious correlations. Therefore, the adversaries can easily fool the model with the manipulation of the pixels with promiscuous information, which includes the various backgrounds, the different viewing perspectives, changing colors, and so on.

\begin{figure}[t]
\centering
\begin{minipage}{.5\textwidth}
  \centering
  \includegraphics[width=.6\linewidth]{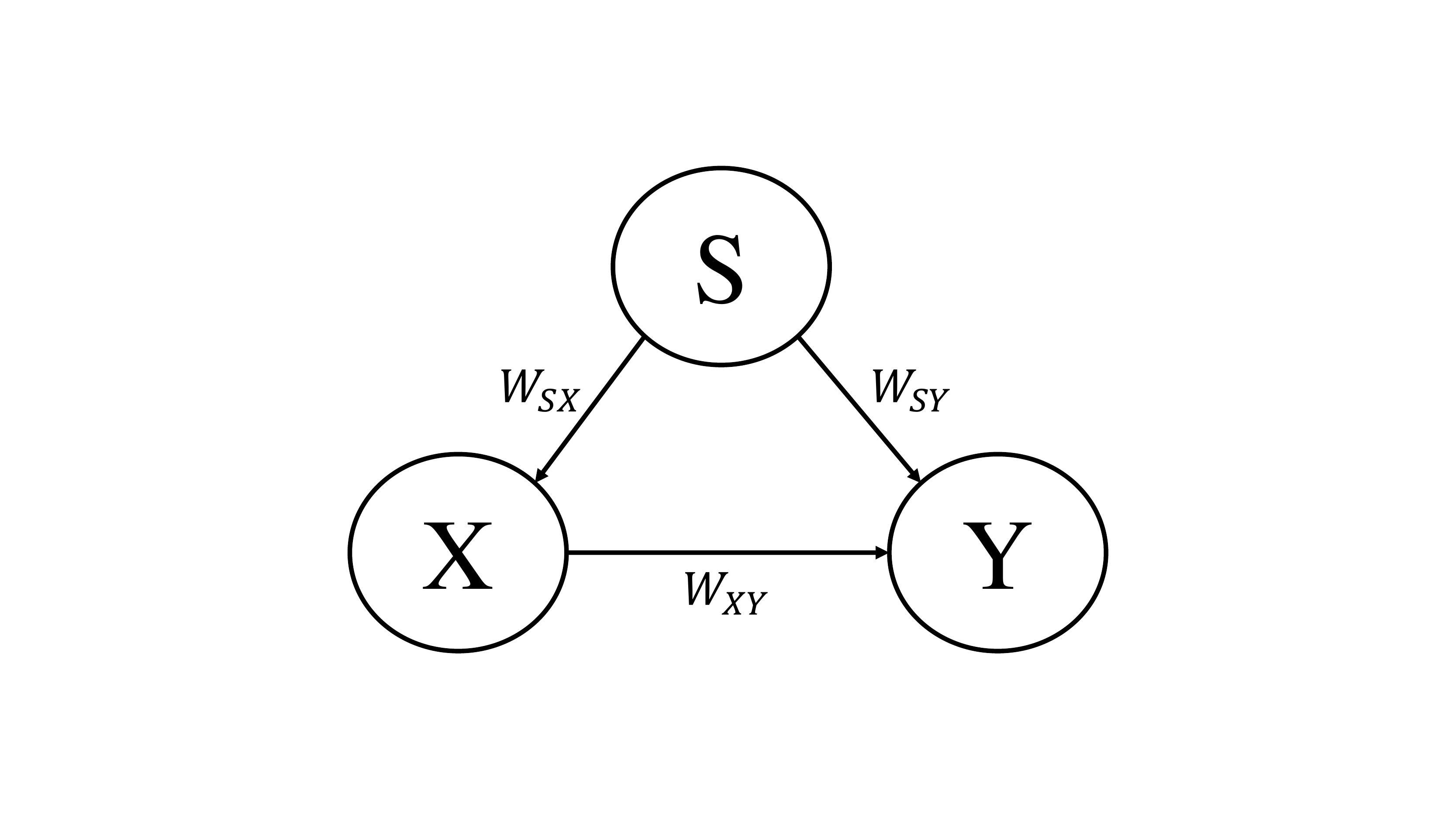}
  \caption{Causal graph without \\
        the instrumental variable.}
  \label{fig:WO-weight}
\end{minipage}%
\begin{minipage}{.5\textwidth}
  \centering
  \includegraphics[width=.6\linewidth]{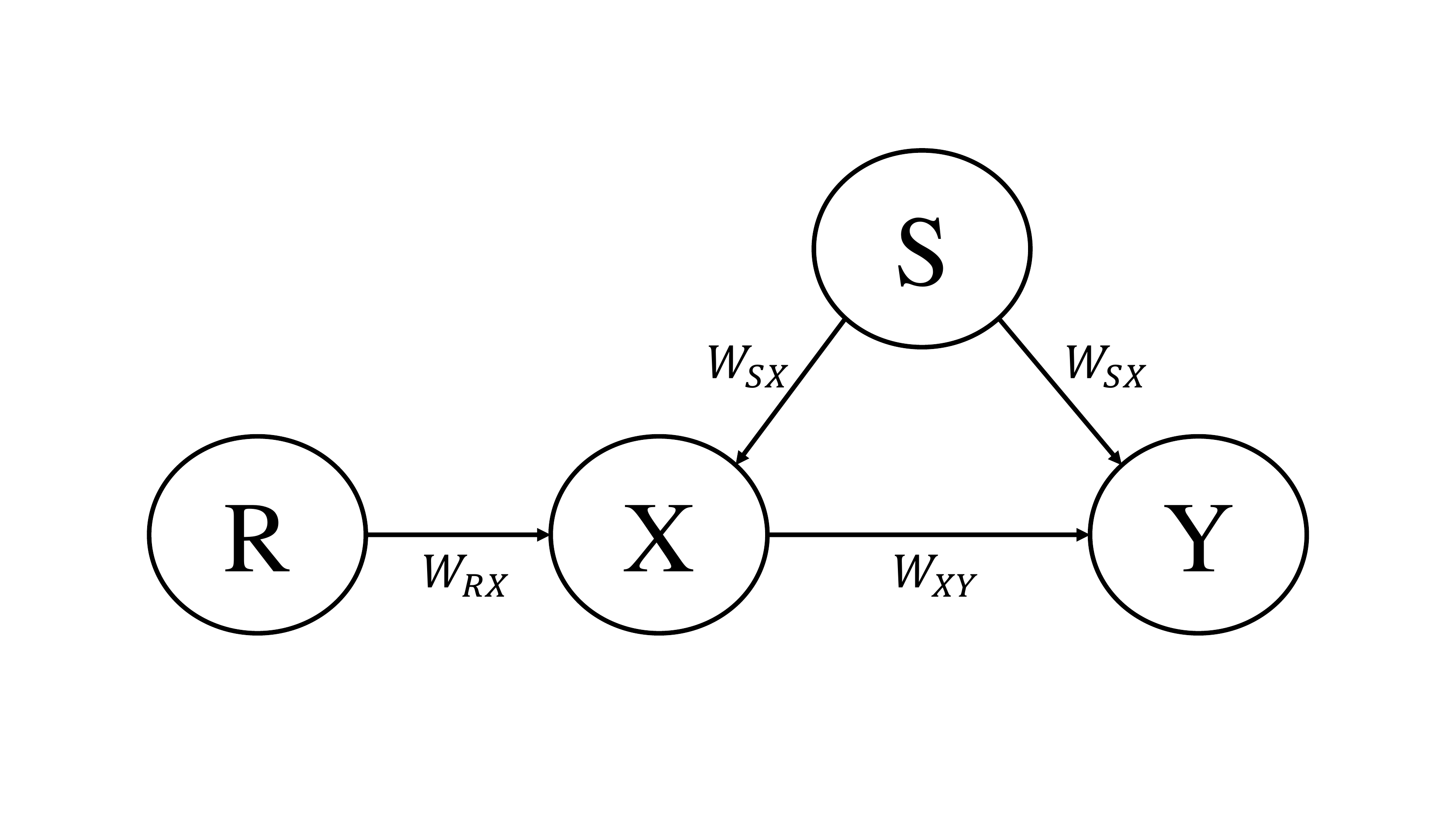}
  \caption{Causal graph with \\
        the instrumental variable.}
  \label{fig:W-weight}
\end{minipage}
\end{figure}

The model focuses more on the robust features rather than the confounding factors if the causal inference methods are incorporated. Based on the causal theory, the generation process of the data can be visualized in Figure \ref{fig:WO-weight}.
% \begin{figure}[htbp]
% \centering
% \includegraphics[width=0.4\textwidth]{WO-R.pdf}
% \caption{FigureTitle}
% \label{fig:woR}
% \end{figure}
A causal graph illustrates the causal relationship between data and features, which would help understand the generation of adversarial examples. In this study, since we have no way to use causal structure learning in complex high-dimensional data, we leverage external knowledge to build the causal graph. As shown in Figure \ref{fig:WO-weight}, the graphical model is implemented where $X$, $Y$, and $S$ represent the original data, prediction, and style information respectively. DNNs should learn robust features which are closely related to the label information. However, the current learning mechanism of DNNs would not distinguish the robust features from non-robust features. $X\gets S\rightarrow Y$ shows that style information is the common cause of $X$ and $Y$, and it affects the distribution of $X$ and $Y$ at the same time. To distinguish the robust features from the non-robust features, the path from $S$ to $Y$ should be truncated. 
%the back-door criterion or the front-door criterion can be used to estimate the causal effect.

% For a pair of variables $(X, Y)$ in a given directed acyclic graph (DAG), if the set of variables $Z$ satisfies: 1) there is no descendant node of $X$ in $S$; 2) $S$ blocks every path between $X$ and $Y$ that contains a node pointing to $X$, then $S$ satisfies the backdoor criterion of $(X, Y)$. If a set of variables $Z$ satisfies the backdoor criterion of $(X, Y)$, then the causal effect of X on Y can be calculated by the following equation:
% \begin{equation}\label{...}
% P(Y=y|do(X=x))=\sum_z{P\left( Y=y|X=x,Z=z \right) P\left( Z=z \right)}.
% \end{equation}
% But due to the lack of sufficient prior knowledge of style information, we do not know the overall distribution of it, Therefore, the backdoor criterion cannot be used to estimate the causal effect of $X$ on $Y$.

% Likewise, if a variable set $Z$: 1) cuts off all paths from $X$ to $Y$; 2) there is no backdoor path from $X$ to $Z$; 3) all backdoor paths from $Z$ to $Y$ are blocked by $X$, then $Z$ satisfies the front door criterion for the variable pair $(X, Y)$, and the causal effect of X on Y can be calculated by the following equation:
% \begin{equation}\label{...}
% P\left( Y=y|do\left( x \right) \right) =\underset{z}{\sum{P\left( z|x \right)}}\sum{P\left( y|x^{\prime},z \right) P\left( x^{\prime} \right)}
% \end{equation}
% However, there is no way we can apply the front door criterion because there is no observable variable set $Z$ between $X$ and $Y$. In this case, 
To obtain the pure causal effect, we turn to instrumental variable estimation. By definition~\cite{CausalityPearl,CausalInference}, a valid instrumental variable should satisfy: 1) it is independent of confounding variables; 2) it affects $Y$ only through $X$. We argue that the artificially introduced additive noise fully meets these two requirements. Figure~\ref{fig:W-weight} illustrates the linear confounded model with the instrumental variable.

When no instrument variable is introduced, $X$ is only affected by $S$, that is, $x=w_{sx}s+u_x$, where  $x$ is the source information, $s$ is the style information, $w_ {sx}$ is the path coefficient between $S$ and $X$, $u_x$ is the independent component of X. And $Y$ is affected by both $X$ and $S$, that is, $y=w_{xy}x+w_{sy}s+u_y$, where $w_{xy} $ is the path coefficient between $X$ and $Y$, $w_{sy}$ is the path coefficient between $S$ and $Y$, $u_y $ is the independent component of $Y$. In this case, we have no way of knowing the causal effect of $X$ on $Y$ because the distribution of $S$ is unobservable.

When the instrument variable $R$ is introduced, $X$ is affected by both $R$ and $S$, that is, $x_r=w_{sx}s+w_{rx}r+u_x $, $y_r=w_{xy}x_r+w_{sy}s+u_y$, where $r$ is the instrument variable, and $w_{rx}$ the path coefficient between $R$ and $X$. To remove the influence of style information $S$, we utilize the direct controlled influence (CDE):
\begin{equation}\label{CDE}
CDE=P\left( Y=y|do\left( X=x_i \right) ,do\left( S=s \right) \right) -P\left( Y=y|do\left( X=x_j \right) ,do\left( S=s \right) \right),
\end{equation}
where $i$ and $j$ are different interventions. Substituting the above formula into Eq.\ref{CDE}, Eq.\ref{...} can be acquired:
\begin{equation}\label{...}
y_{r_i}-y_{r_j}=w_{xy}\left( x_{r_i}-x_{r_j} \right),
\end{equation}					
then the robust feature $w_{xy}$ can be learned by Eq.\ref{wxy}:
\begin{equation}\label{wxy}
w_{xy}=\frac{y_{r_i}-y_{r_j}}{x_{r_i}-x_{r_j}}.
\end{equation}
\subsection{The Proposed CausalIB}
In practice, we refer to the idea of CiiV~\cite{CiiV}, which uses the retinal mask as the instrumental variable in the original image layer to obtain images under different gaze angles, and achieves good results. However, due to the property of representation learning in DNNs, we postulate that it would be more rational to use instrumental variables in the intermediate feature level instead of the original image layer. Therefore, we introduce additive Gaussian noise as the instrumental variable in the feature layer. Next, we will specifically describe the proposed method.

Our method is based on VIB~\cite{VIB}, which learns multi-dimensional Gaussian distributions instead of deterministic features in feature layers. VIB takes the mean of samples of the feature distribution, but the difference between these feature layer samples is neglected. Such a difference can be modeled via the additive noises as the instrumental variable. In our framework, the data distribution rather than the final value of the mean output was subjected to causal inference. 

The relationship between $Y$ and $R$ can be written as Eq.\ref{causalEquation}:
\begin{equation}\label{causalEquation}
Y[X=x_r]=w_{xy}x_r+w_{sy}s=w_{sy}s+w_{xy}w_{sx}s+w_{rx}w_{xy}r=Y[X=x]+w_{rx}w_{xy}r.
\end{equation}
Therefore, the Eq.\ref{wxy} can be written as Eq.\ref{causalEffect}:
\begin{equation}\label{causalEffect}
w_{xy_i}=w_{rx}w_{xy}=\frac{Y\left[ X=x_{r_i} \right] -Y\left[ X=x \right]}{r_i},
\end{equation}
where $x_{r_i}$ is the $i$th sample of $X$ with the instrumental variable. We assume that the magnitude of $r$ is equal to the intensity of the introduced noise. In practice, since $R$ is independent of $S$, the causal loss function can be formalized as Eq.\ref{LossCausal}:
\begin{equation}\label{LossCausal}
\mathcal{L} _{causal}=\sum_{i\ne j}{||w_{xy_i}-w_{xy_j}||}.
\end{equation}
Combined with the IB loss function, the proposed CausalIB loss function can be expressed as Eq.\ref{totalLoss}:
\begin{equation}
\mathcal{L}_{CausalIB}=\mathcal{L}_{IB}+\mathcal{L}_{causal}
\\
\,\, =-I\left( Z;Y \right) +\alpha I\left( Z;X \right) +\beta\sum_{i\ne j}{||w_{xy_i}-w_{xy_j}||}. \label{totalLoss}
\end{equation}
Given $X$, Eq.\ref{totalLoss} encourages the model to learn $w_{xy}$ undisturbed by $w_{sy}$, and the learned features can be generally compressed. $\alpha$ and $\beta$ are hyperparameters.
\section{Experiments}
In this section, we verify the efficacy of the proposed CausalIB method by numerical experiments.
\subsection{Datasets and Settings}
\subsubsection{Datasets}
We apply the proposed CausalIB model on three benchmark datasets (MNIST~\cite{MNISTref}, FashionMNIST~\cite{fashionMNISTref}, and CIFAR-10~\cite{krizhevsky2009learning}) and evaluate its adversarial robustness. MNIST and FashionMNIST contain 65K handwritten digit image samples and commodity image samples respectively, and the size is 28x28. CIFAR-10 contains 60K classified image samples with a size of 32x32.
\subsubsection{Training Details}

\begin{table}[!t]
\caption{The performances of white-box attack on MNIST and FashionMNIST.}
\centering
\scriptsize

\begin{tabular}{c|ccc|ccc}
\hline
\multicolumn{1}{l|}{}              & \multicolumn{3}{c|}{MNIST}                        & \multicolumn{3}{c}{FashionMNIST}                 \\ \hline
\textbf{Method} & \textbf{clean} & \textbf{FGSM} & \textbf{PGD-20} & \textbf{clean} & \textbf{FGSM} & \textbf{PGD-20} \\ \hline
Baseline        & 95.11\thinspace\thinspace$\pm$\thinspace \thinspace 0.72 & 6.76\thinspace$\pm$\thinspace 0.93  & 0.0\thinspace$\pm$\thinspace 0.0  & 89.81\thinspace$\pm$\thinspace 0.49 & 0.63\thinspace$\pm$\thinspace 0.34  & 0.0\thinspace$\pm$\thinspace 0.0    \\
mixu  & 98.22\thinspace$\pm$\thinspace 0.81 & 20.66\thinspace$\pm$\thinspace 1.51 & 0.43\thinspace$\pm$\thinspace 0.08  & 90.44\thinspace$\pm$\thinspace 0.29 & 9.13\thinspace$\pm$\thinspace 3.05  & 0.11\thinspace$\pm$\thinspace 0.04  \\
RS     & 98.45\thinspace$\pm$\thinspace 0.53 & 34.15\thinspace$\pm$\thinspace 0.91 & 13.38\thinspace$\pm$\thinspace 0.93 & 89.71\thinspace$\pm$\thinspace 0.22 & 13.57\thinspace$\pm$\thinspace 1.13 & 1.30\thinspace$\pm$\thinspace 0.21  \\
CiiV            & 97.94\thinspace$\pm$\thinspace 0.06 & 71.80\thinspace$\pm$\thinspace 0.32 & 48.48\thinspace$\pm$\thinspace 0.77 & \pmb{94.84\thinspace$\pm$\thinspace 0.15} & 44.82\thinspace$\pm$\thinspace 0.87 & \pmb{15.01\thinspace$\pm$\thinspace 0.34} \\
VIB             & 98.70\thinspace$\pm$\thinspace 0.26 & 66.06\thinspace$\pm$\thinspace 0.21 & 41.84\thinspace$\pm$\thinspace 0.33 & 94.42\thinspace$\pm$\thinspace 0.44 & 25.88\thinspace$\pm$\thinspace 0.41 & 3.98\thinspace$\pm$\thinspace 0.58  \\
CausalIB        & \pmb{98.85\thinspace$\pm$\thinspace 0.20} & \pmb{81.07\thinspace$\pm$\thinspace 0.82} & \pmb{53.71\thinspace$\pm$\thinspace 0.51} & 94.06\thinspace$\pm$\thinspace 0.25 & \pmb{50.81\thinspace$\pm$\thinspace 0.80} & 14.06\thinspace$\pm$\thinspace 0.11 \\
AT (FGSM)       & 94.02\thinspace$\pm$\thinspace 0.12 & 75.82\thinspace$\pm$\thinspace 1.13 & 67.86\thinspace$\pm$\thinspace 2.35 & 89.95\thinspace$\pm$\thinspace 0.15 & 55.52\thinspace$\pm$\thinspace 0.80 & 17.46\thinspace$\pm$\thinspace 1.76 \\
AT (PGD-20)      & 93.92\thinspace$\pm$\thinspace 0.05 & 78.31\thinspace$\pm$\thinspace 2.58 & 69.84\thinspace$\pm$\thinspace 0.94 & 87.22\thinspace$\pm$\thinspace 0.49 & 56.43\thinspace$\pm$\thinspace 0.61 & 41.87\thinspace$\pm$\thinspace 1.60 \\ \hline
\end{tabular}
\label{MNISTandFashionMNISTLable}
\end{table}

% \begin{table}
% \caption{The performances of white-box attack on MNIST.}
% \label{MNISTTable}
%     \centering
%     \begin{tabular}{lllllllll}  
%     \cmidrule(r){1-4}
%     Method         & clean       & FGSM        & PGD-20      \\
%     \cmidrule(r){1-4}
%     Baseline       & 95.11\thinspace\thinspace$\pm$\thinspace \thinspace 0.72 & 6.76\thinspace$\pm$\thinspace 0.93  & 0.0\thinspace$\pm$\thinspace 0.0    \\
%     mixup~\cite{mixup}         & 98.22\thinspace$\pm$\thinspace 0.81 & 20.66\thinspace$\pm$\thinspace 1.51 & 0.43\thinspace$\pm$\thinspace 0.08  \\
%     RS~\cite{RS}             & 98.45\thinspace$\pm$\thinspace 0.53 & 34.15\thinspace$\pm$\thinspace 0.91 & 13.38\thinspace$\pm$\thinspace 0.93 \\
%     CiiV~\cite{CiiV}         & 97.94\thinspace$\pm$\thinspace 0.06 & 71.80\thinspace$\pm$\thinspace 0.32 & 48.48\thinspace$\pm$\thinspace 0.77 \\
%     VIB~\cite{VIB}            & 98.70\thinspace$\pm$\thinspace 0.26 & 66.06\thinspace$\pm$\thinspace 0.21 & 41.84\thinspace$\pm$\thinspace 0.33 \\
%     CausalIB      & \pmb{98.85\thinspace$\pm$\thinspace 0.20} & \pmb{81.07\thinspace$\pm$\thinspace 0.82} & \pmb{53.71\thinspace$\pm$\thinspace 0.51} \\
%     AT (FGSM)~\cite{TowardsAdver}   & 94.02\thinspace$\pm$\thinspace 0.12 & 75.82\thinspace$\pm$\thinspace 1.13 & 67.86\thinspace$\pm$\thinspace 2.35 \\
%     AT (PGD-20)~\cite{TowardsAdver}  & 93.92\thinspace$\pm$\thinspace 0.05 & 78.31\thinspace$\pm$\thinspace 2.58 & 69.84\thinspace$\pm$\thinspace 0.94
    
%     \end{tabular}
    
% \end{table}
The experiments in this study are carried out on Tesla P100 and tested statistically. Since our purpose is to study the IB method and the improvement of model robustness by causal inference, complex models are not chosen. In the experiments of MNIST and FashionMNIST, our basic model is a simple three-layer MLP, and the CausalIB method is applied in the last layer. In the experiments on the CIFAR-10 dataset, our basic model is AlexNet~\cite{alexnet}, which also uses the CausalIB method in the last layer. All models are trained using the Adam optimizer, 100 samples per batch, and 50 epochs per training.
\subsubsection{Details of  Threat Models}
The attack models in this study are FGSM, and PGD-20. In the MNIST and FashionMNIST experiments, we choose the budget radius $\epsilon$ to be 50/255. In the CIFAR-10 experiment, we choose the budget radius $\epsilon$ to be 8/255.
\subsubsection{Details of other Defense Models}
For adversarial training methods, we adopted two popular defenders: AT (FGSM) and AT (PGD-20)~\cite{TowardsAdver}, and these two methods are implemented with the same FGSM and PGD-20 parameters as the experiments. For the methods without the utilization of adversarial training, we investigated mixup~\cite{mixup}, randomized smoothing (RS)~\cite{RS}, VIB~\cite{VIB} and CiiV~\cite{CiiV}. Mixup is a data augmentation method that uses mixed sample data augmentation to mix images between different classes to augment the training dataset. RS uses smoothing any function into a gradient-bounded function to improve the robustness of the 
model, and mathematically it is well proven. VIB utilizes variational inference methods to optimize information bottlenecks and utilizes the reparameterization trick for efficient training. It has achieved the considerable results in improving model generalization and adversarial attack robustness. CiiV is a causal inference-based model, which augments the image with multiple retinal subject centers, encouraging the model to learn causal features, rather than local confusion patterns. Also, it can be combined with other methods to achieve the considerable results in improving the robustness of the model.
\subsection{Robustness evaluation}
We report the comparison of our CausalIB method and other methods without the utilization of adversarial training in Table \ref{MNISTandFashionMNISTLable} and Table \ref{CIFARTable}, where the "clean" item in the tables represents the classification accuracy of the model on clean examples. We also provide adversarial training results at the bottom of each table for comparison. It can be seen that the proposed CausalIB shows the best overall performance in all methods without the utilization of adversarial training, demonstrating that taking into account the spurious correlation can significantly improve the adversarial robustness. In Table \ref{MNISTandFashionMNISTLable}, the CausalIB shows complete superiority on MNIST, and even outperforms adversarial training methods in adversarial defense against FGSM. However, the performance of methods without the utilization of adversarial training is inferior to the performance of adversarial training on complex datasets such as CIFAR-10, which can be seen in Table \ref{CIFARTable}. The advantage of the methods without the utilization of adversarial training is that they guarantee the classification accuracy of clean examples. In the experiment on FashionMNIST, the classification accuracy of clean examples of CausalIB is about 7\% higher than that of adversarial training methods.
\begin{figure}[t]
\centering
\begin{subfigure}{0.48\textwidth}
\includegraphics[width=2.1in, height=1.5in]{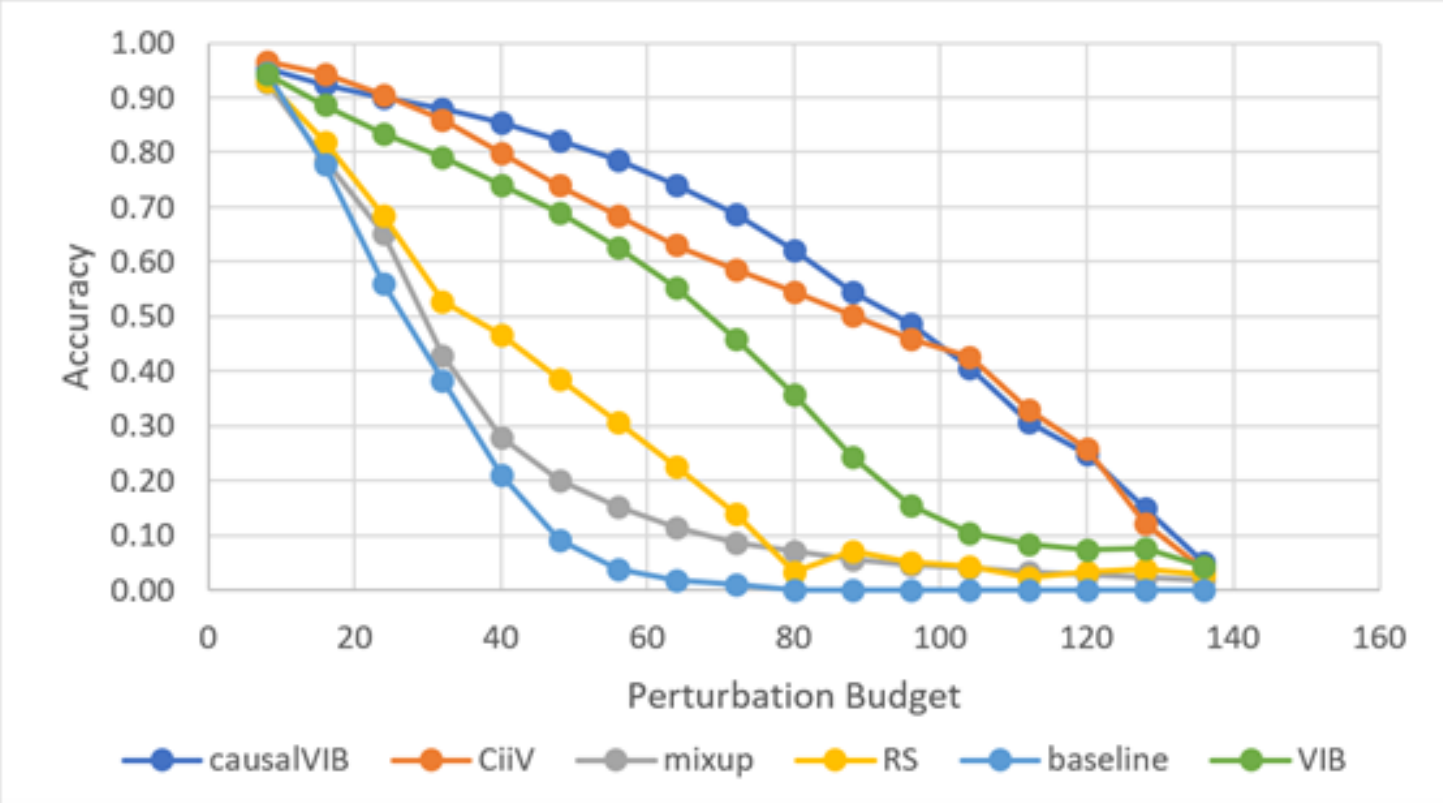} 
\caption{FGSM}
\label{fig:subim1}
\end{subfigure}
\begin{subfigure}{0.48\textwidth}
\includegraphics[width=2.1in, height=1.5in]{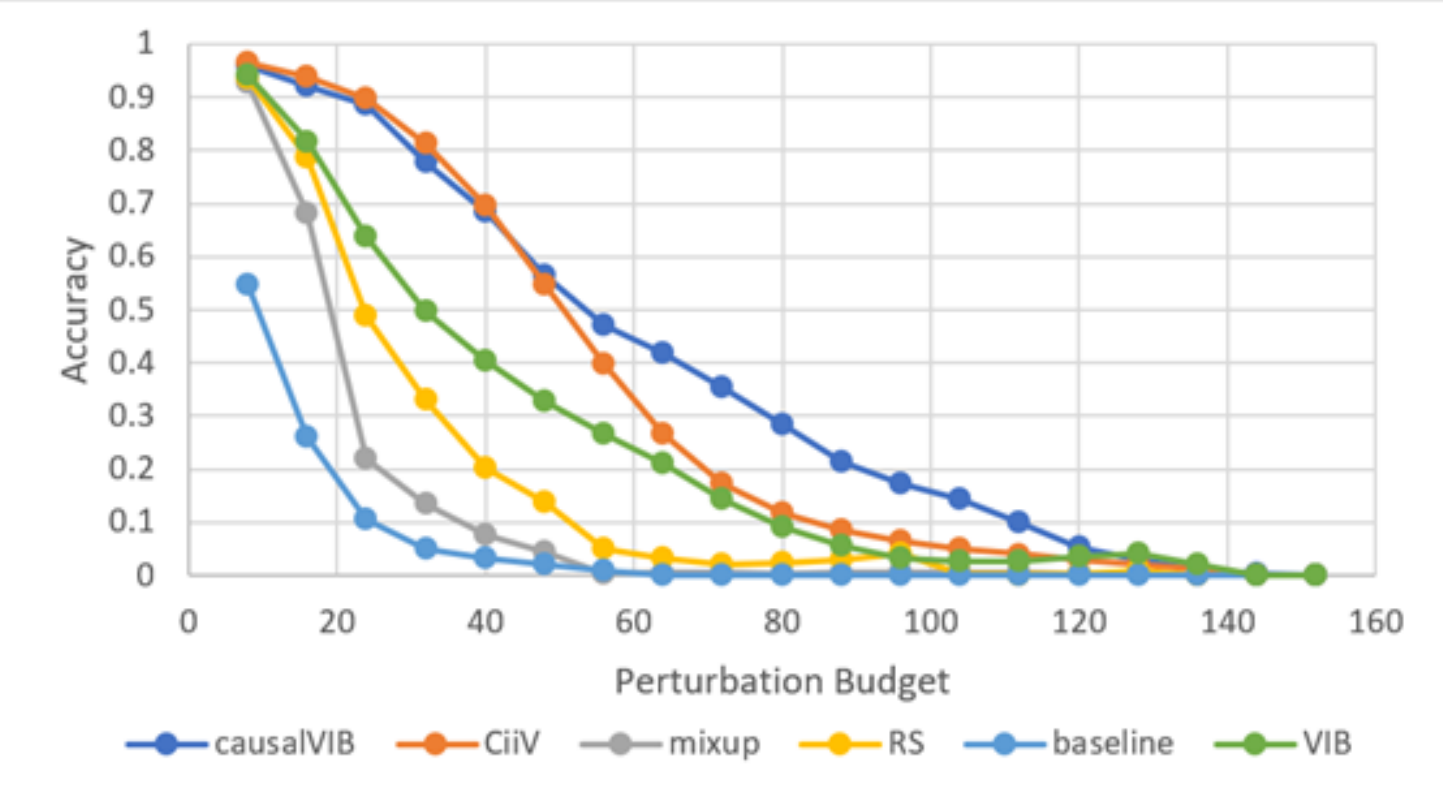}
\caption{PGD}
\label{fig:subim2}
\end{subfigure}
\caption{Unbounded attacks on MNIST that increase the budget radius $\epsilon$ from 8/255 to 152/255.}
\label{fig:image2}
\end{figure}

\subsection{Adversarial Robustness Under Unbounded Attack}
To evaluate the validity of defenders, we compare the performances of CausalIB with other methods without the utilization of adversarial training under unbounded attacking in Figure \ref{fig:image2}. Under the same budget radius $\epsilon$, our method is better than VIB. Under the single-step attack, the classification accuracy of VIB drops below 50\% at 72/255, while the classification accuracy of CausalIB drops below 50\% at 96/255. Similarly, in the case of iterative attack, the performance of CausalIB is about 20\% better than VIB and maintains a stable rate of descent at the high budget radius. When the budget radius $\epsilon$ of the attacker was increased from 8/255 to 152/255, all performances were converged to the random guesses or even worse. Any valid defender shouldn’t survive such an unbounded attack, as it allows the attacker to modify the entire image and erase all causal features.
\begin{table}[!t]
\caption{The performances of white-box attack and black-box attack on CIFAR-10.}
\centering
\scriptsize
\begin{tabular}{c|ccc|cc}
\hline
\multicolumn{1}{l|}{} & \multicolumn{3}{c|}{White-box}           & \multicolumn{2}{c}{Black-box} \\ \hline
Method                & clean       & FGSM        & PGD-20      & FGSM          & PGD-20       \\ \hline
Baseline~\cite{alexnet}       & 88.06\thinspace$\pm$\thinspace 0.87 & 30.48\thinspace$\pm$\thinspace 0.36 & 0.12\thinspace$\pm$\thinspace 0.13  & 31.60\thinspace$\pm$\thinspace 1.12 & 0.52\thinspace$\pm$\thinspace 1.26  \\
mixup~\cite{mixup}          & 88.46\thinspace$\pm$\thinspace 1.23 & 42.54\thinspace$\pm$\thinspace 1.57 & 12.33\thinspace$\pm$\thinspace 1.98 & 42.54\thinspace$\pm$\thinspace 1.06 & 12.13\thinspace$\pm$\thinspace 0.96 \\
RS~\cite{RS}             & 90.40\thinspace$\pm$\thinspace 0.53 & 44.22\thinspace$\pm$\thinspace 0.94 & 22.23\thinspace$\pm$\thinspace 1.02 & 49.23\thinspace$\pm$\thinspace 1.42 & 28.23\thinspace$\pm$\thinspace 1.46 \\
CiiV~\cite{CiiV}           & 89.50\thinspace$\pm$\thinspace 0.87 & 52.95\thinspace$\pm$\thinspace 0.43 & 33.62\thinspace$\pm$\thinspace 0.78 & 53.95\thinspace$\pm$\thinspace 0.73 & 34.85\thinspace$\pm$\thinspace 0.82 \\
VIB~\cite{VIB}            & \pmb{91.93\thinspace$\pm$\thinspace 0.17} & 41.89\thinspace$\pm$\thinspace 0.98 & 22.81\thinspace$\pm$\thinspace 1.36 & 39.40\thinspace$\pm$\thinspace 0.46 & 23.31\thinspace$\pm$\thinspace 0.77 \\
CausalIB      & 91.76\thinspace$\pm$\thinspace 0.25 & \pmb{54.11\thinspace$\pm$\thinspace 0.61} & \pmb{35.89\thinspace$\pm$\thinspace 0.34} & 56.36\thinspace$\pm$\thinspace 1.28 & 36.02\thinspace$\pm$\thinspace 1.51 \\
AT (FGSM)~\cite{TowardsAdver}   & 85.49\thinspace$\pm$\thinspace 0.62 & 86.32\thinspace$\pm$\thinspace 0.75 & 23.40\thinspace$\pm$\thinspace 0.93 & - & - \\
AT (PGD-20)~\cite{TowardsAdver} & 85.39\thinspace$\pm$\thinspace 0.86 & 65.91\thinspace$\pm$\thinspace 1.02 & 61.20\thinspace$\pm$\thinspace 0.67  & - & - \\ \hline
\end{tabular}
\label{CIFARTable}
\end{table}
%\subsection{Visualization}
%We make a visual comparison of VIB and CausalIB in the appendix of supplementary material. The adversarial examples of VIB are relatively blurry, while the adversarial examples of CausalIB have much clearer structural details. This means that the causal inference method could lead the model to learn more structural information. As a result, the adversary must add larger perturbations and try to erase the structural patterns to fool the CausalIB model. 
% \begin{figure}[ht]
% \begin{subfigure}{0.3\textwidth}
% \centering
% \includegraphics[width=1.5in, height=2.0in]{littleadv.png} 
% \caption{FGSM}
% \label{fig:subim1}
% \end{subfigure}
% \begin{subfigure}{0.7\textwidth}
% \centering
% \includegraphics[width=4.0in, height=2.0in]{adv.png}
% \caption{PGD}
% \label{fig:subim2}
% \end{subfigure}
% \caption{Visualization of adversarial examples of VIB and CausalIB}
% \label{fig:image2}
% \end{figure}
\subsection{Ablation Studies}
\begin{table}[!t]
\centering
\scriptsize
\caption{Ablation experiments on MNIST.}
\label{Ablation}
\begin{tabular}{ccccc}
\cmidrule(r){1-4}
\textbf{Method}       & \textbf{clean} & \textbf{FGSM} & \textbf{PGD-20} &  \\
\cmidrule(r){1-4}
$L_1$                  & 98.85\thinspace$\pm$\thinspace 0.20    & 81.07\thinspace$\pm$\thinspace 0.82   & 53.71\thinspace$\pm$\thinspace 0.51     &  \\
$L_2$                  & 97.37\thinspace$\pm$\thinspace 0.53    & 78.55\thinspace$\pm$\thinspace 0.93   & 45.62\thinspace$\pm$\thinspace 1.21     &  \\
$\alpha$=0/ $\beta$=0       & 95.11\thinspace$\pm$\thinspace 0.72    & 6.76\thinspace$\pm$\thinspace 0.93    & 0.0\thinspace$\pm$\thinspace 0.0        &  \\
$\alpha$=0.05/ $\beta$=0    & 98.41\thinspace$\pm$\thinspace 0.64    & 63.61\thinspace$\pm$\thinspace 1.49   & 38.98\thinspace$\pm$\thinspace 1.23     &  \\
$\alpha$=0.05/ $\beta$=0.05  & 98.85\thinspace$\pm$\thinspace 0.20    & 81.07\thinspace$\pm$\thinspace 0.82   & 53.71\thinspace$\pm$\thinspace 0.51     &  \\
$\alpha$=0.05/ $\beta$=1.0  & 92.69\thinspace$\pm$\thinspace 1.11    & 68.32\thinspace$\pm$\thinspace 1.54   & 20.49\thinspace$\pm$\thinspace 1.84     &  \\
$\alpha$=0.01/ $\beta$=0.05 & 95.62\thinspace$\pm$\thinspace 0.34    & 74.86\thinspace$\pm$\thinspace 0.64   & 29.65\thinspace$\pm$\thinspace 1.41     &  
\end{tabular}
\end{table}
In this section, we evaluate the performance of the proposed CausalIB under different settings and parameters on the MNIST dataset. 1) As shown in Table \ref{Ablation}, the loss metrics with different norms are studied, where $L_1$ loss is better than $L_2$ loss;
2) Other choices of hyperparameters of the CausalIB method are reported. It can be found that $\beta$ is used as a trade-off between the classification accuracy of clean examples and the performance on adversarial examples. A larger $\beta$ will lead to a larger drop in the model's performance on clean examples, and when $\beta$ is too large, the adversarial defense performance will also drop. In practice, the setting of hyperparameters depends on empirical experiments. 3) Our model has two parameters that need to be weighed. The IB parameter $\alpha$ used in our method is different from VIB. In experiments of VIB, the model performs best when $\alpha$ is 0.01, while the $\alpha$ in the CausalIB method is set as 0.05, we also provide relevant experimental results. It can be seen that causal inference further improves the robustness of the model to adversarial examples based on the
IB method.
\subsection{Experiments with Auto-Attack Threat Model}
\textcolor{black}{In this section, more detailed experiment results of the performances of Auto-Attack~\cite{autoattack} threat model with the budget radius $\epsilon = 8/255$ are provided. AutoAttack is a threat model that integrates various parameterless attacks, including APGD-CE \cite{autoattack}, APGD-DLR \cite{autoattack}, the black-box Square Attack \cite{blacksqure}, and the FAB attack \cite{FAB}. VIB~\cite{VIB}, CiiV~\cite{CiiV}, and our proposed method are evaluated on MNIST, FashionMNIST, CIFAR-10, and CIFAR-100. MLPs are evaluated structures on MNIST and FashionMNIST, and AlexNet~\cite{alexnet} is evaluated on CIFAR-10. On CIFAR-100, ResNet34 with or without regularization methods are evaluated under the Auto-Attack threat. As shown in Table \ref{tableM} and Table \ref{tableC}, the proposed CausalIB shows the overall considerable defense performance under the Auto-Attack threat.}
\begin{table}[!t]
\centering
\scriptsize
\caption{The performances of Auto-Attack on MNIST and FashionMNIST.}
\begin{tabular}{ccccccc}
\hline
         & \multicolumn{3}{c}{\textbf{MNIST}}                                                & \multicolumn{3}{c}{\textbf{FashionMNIST}}\\
\hline
                     & clean          & AA-$L_{inf}$       & AA-$L_2$          & clean & AA-$L_{inf}$        & AA-$L_2$          \\
\hline
MLP w/o defense            & 95.11          & 56.01         & 67.19          & 89.81 & 34.16          & 48.92          \\
VIB~\cite{VIB}                  & 98.70          & 89.01         & 92.68          & 94.42 & 65.21          & 76.54          \\
CiiV~\cite{CiiV}                 & 97.94          & 94.88         & 96.03          & \textbf{94.84}  & 78.42          & 83.03          \\
causalIB             & \textbf{98.85} & \textbf{95.10} & \textbf{96.89} & 94.06 & \textbf{80.77} & \textbf{83.72}     
\end{tabular}
\label{tableM}
\end{table}

\begin{table}[!t]
\centering
\scriptsize
\caption{The performances of Auto-Attack on CIFAR-10 and CIFAR-100.}
\begin{tabular}{ccccccc}
\hline
         & \multicolumn{3}{c}{\textbf{CIFAR-10}}                    & \multicolumn{3}{c}{\textbf{CIFAR-100}}                   \\
\hline
         & clean          & AA-$L_{inf}$       & AA-$L_2$          & clean         & AA-$L_{inf}$        & AA-$L_2$          \\
\hline
CNNs w/o defense~\cite{alexnet,ResNet}  & 88.06          & 0.0           & 0.0            & \textbf{70.1} & 0.0            & 0.0            \\
VIB~\cite{VIB}      & 91.93          & 18.60          & 63.58          & 56.48         & 8.49           & 36.80           \\
CiiV~\cite{CiiV}     & 89.50          & 26.33         & \textbf{70.24} & 52.15         & 18.34          & 43.36          \\
causalIB & \textbf{91.76} & \textbf{28.7} & 69.93          & 52.58         & \textbf{20.12} & \textbf{43.89} 
\end{tabular}
\label{tableC}
\end{table}

\subsection{Gradient Obfuscation Discussion}
To verify that the proposed CausalIB does not suffer from flawed or incomplete evaluations, our experiments were designed to follow a series of sanity checks:
\begin{compactenum}
\item The experimental results on FGSM are better than PGD-20. 
\item The experimental results on black-box attacks are better than on white-box attacks. 
\item The result of a weak attack (FGSM) is better than that of a strong attack (PGD-20). 
\item Unbounded adversarial examples become random guessing or 0\% accuracy.
\end{compactenum}
These four phenomena tesify that the proposed CausalIB does not suffer from the problem of gradient obfuscation~\cite{evaluateLaw}. The experiment of the defense against adaptive attack method~\cite{TowardsRoubustNeural} would be illustrated in the appendix.
\subsection{Shortcomings of the CausalIB method}
Although our proposed CausalIB has improved adversarial defense performance, it still has the following shortcomings:
\begin{compactenum}
\item The causal inference method leads to a further decrease in cleaning performance relative to VIB in complex datasets due to the addition of another trade-off, which is reported in the empirical results on the CIFAR-10 dataset.
\item The settings of hyperparameters are empirical.
\item The linear causal model cannot reflect all the scenarios accurately, because DNNs are highly nonlinear. In our future work, a more complex causal graph should be modeled.
\item Currently, the adversarial training methods based on min-max optimization are state-of-the-art (SOTA) defense methods. The adversarial examples are the training samples in the optimization process of adversarial training. In contrast, the generation of adversarial examples is not essential in the training process of CausalIB. Our empirical study on CIFAR-10 also validates the performance gap between CausalIB and AT (PGD-20). In the future, it would be promising to combine the CausalIB with the framework of adversarial training.
\end{compactenum}
\section{Conclusion}
In this paper, we address the inadequacies of the spurious correlation in the IB framework that hinder adversarial robustness. We analyze the causes of adversarial examples with the utilization of a causal graph to demonstrate that the spurious correlations between robust features and non-robust features are one of the problems to be alleviated in current IB methods. In the proposed CausalIB method, additive noises are used as an instrumental variable to estimate the causal effect. Such a method can separate robust features and non-robust features. The utilization of this method improves the performance of the IB method in adversarial defense to a certain extent. In future work, we will study whether the CausalIB method can be applied in other scenarios such as OOD image classification.

\bibliography{iclr2023_conference}

\begin{thebibliography}{69}
\providecommand{\natexlab}[1]{#1}
\providecommand{\url}[1]{\texttt{#1}}
\expandafter\ifx\csname urlstyle\endcsname\relax
  \providecommand{\doi}[1]{doi: #1}\else
  \providecommand{\doi}{doi: \begingroup \urlstyle{rm}\Url}\fi

\bibitem[Achille \& Soatto(2018)Achille and Soatto]{InformationDropout}
Alessandro Achille and Stefano Soatto.
\newblock Information dropout: Learning optimal representations through noisy
  computation.
\newblock \emph{IEEE}, 40\penalty0 (12):\penalty0 2897--2905, 2018.

\bibitem[Akhtar \& Mian(2018)Akhtar and Mian]{ThratOfAdverAttack}
Naveed Akhtar and Ajmal~S. Mian.
\newblock Threat of adversarial attacks on deep learning in computer vision:
  {A} survey.
\newblock \emph{{IEEE} Access}, 6:\penalty0 14410--14430, 2018.

\bibitem[Akhtar et~al.(2018)Akhtar, Liu, and Mian]{akhtar2018defense}
Naveed Akhtar, Jian Liu, and Ajmal Mian.
\newblock Defense against universal adversarial perturbations.
\newblock In \emph{CVPR}, pp.\  3389--3398, 2018.

\bibitem[Alemi et~al.(2017)Alemi, Fischer, Dillon, and Murphy]{VIB}
Alexander~A. Alemi, Ian Fischer, Joshua~V. Dillon, and Kevin Murphy.
\newblock Deep variational information bottleneck.
\newblock In \emph{ICLR}, 2017.

\bibitem[Andriushchenko et~al.(2020)Andriushchenko, Croce, Flammarion, and
  Hein]{blacksqure}
Maksym Andriushchenko, Francesco Croce, Nicolas Flammarion, and Matthias Hein.
\newblock Square attack: {A} query-efficient black-box adversarial attack via
  random search.
\newblock volume 12368, pp.\  484--501, 2020.

\bibitem[Bai et~al.(2017)Bai, Quan, and Luo]{bai2017alleviating}
Wenjun Bai, Changqin Quan, and Zhiwei Luo.
\newblock Alleviating adversarial attacks via convolutional autoencoder.
\newblock In Teruhisa Hochin, Hiroaki Hirata, and Hiroki Nomiya (eds.),
  \emph{IEEE/ACIS}, pp.\  53--58, 2017.

\bibitem[Brown et~al.(1990)Brown, Cocke, Pietra, Pietra, Jelinek, Lafferty,
  Mercer, and Roossin]{machineTranslation}
Peter~F. Brown, John Cocke, Stephen~Della Pietra, Vincent J.~Della Pietra,
  Frederick Jelinek, John~D. Lafferty, Robert~L. Mercer, and Paul~S. Roossin.
\newblock A statistical approach to machine translation.
\newblock \emph{Comput. Linguistics}, 16\penalty0 (2):\penalty0 79--85, 1990.

\bibitem[Carlini \& Wagner(2017{\natexlab{a}})Carlini and Wagner]{CW}
Nicholas Carlini and David~A. Wagner.
\newblock Towards evaluating the robustness of neural networks.
\newblock In \emph{{IEEE} Symposium on Security and Privacy (SP)}, pp.\
  39--57. {IEEE} Computer Society, 2017{\natexlab{a}}.

\bibitem[Carlini \& Wagner(2017{\natexlab{b}})Carlini and
  Wagner]{TowardsRoubustNeural}
Nicholas Carlini and David~A. Wagner.
\newblock Towards evaluating the robustness of neural networks.
\newblock In \emph{SP}, pp.\  39--57, 2017{\natexlab{b}}.

\bibitem[Carlini et~al.(2019)Carlini, Athalye, Papernot, Brendel, Rauber,
  Tsipras, Goodfellow, Madry, and Kurakin]{evaluateLaw}
Nicholas Carlini, Anish Athalye, Nicolas Papernot, Wieland Brendel, Jonas
  Rauber, Dimitris Tsipras, Ian~J. Goodfellow, Aleksander Madry, and Alexey
  Kurakin.
\newblock On evaluating adversarial robustness.
\newblock \emph{CoRR}, abs/1902.06705, 2019.

\bibitem[Cohen et~al.(2019)Cohen, Rosenfeld, and Kolter]{RS}
Jeremy~M. Cohen, Elan Rosenfeld, and J.~Zico Kolter.
\newblock Certified adversarial robustness via randomized smoothing.
\newblock In \emph{ICML}, pp.\  1310--1320, 2019.

\bibitem[Croce \& Hein(2020{\natexlab{a}})Croce and Hein]{FAB}
Francesco Croce and Matthias Hein.
\newblock Minimally distorted adversarial examples with a fast adaptive
  boundary attack.
\newblock pp.\  2196--2205, 2020{\natexlab{a}}.

\bibitem[Croce \& Hein(2020{\natexlab{b}})Croce and Hein]{autoattack}
Francesco Croce and Matthias Hein.
\newblock Reliable evaluation of adversarial robustness with an ensemble of
  diverse parameter-free attacks.
\newblock volume 119, pp.\  2206--2216, 2020{\natexlab{b}}.

\bibitem[Das et~al.(2017)Das, Shanbhogue, Chen, Hohman, Chen, Kounavis, and
  Chau]{KeepBadGuysOut}
Nilaksh Das, Madhuri Shanbhogue, Shang{-}Tse Chen, Fred Hohman, Li~Chen,
  Michael~E. Kounavis, and Duen~Horng Chau.
\newblock Keeping the bad guys out: Protecting and vaccinating deep learning
  with {JPEG} compression.
\newblock \emph{CoRR}, abs/1705.02900, 2017.

\bibitem[Devlin et~al.(2018)Devlin, Chang, Lee, and Toutanova]{BERT}
Jacob Devlin, Ming{-}Wei Chang, Kenton Lee, and Kristina Toutanova.
\newblock {BERT:} pre-training of deep bidirectional transformers for language
  understanding.
\newblock \emph{CoRR}, abs/1810.04805, 2018.

\bibitem[Eykholt et~al.(2018)Eykholt, Evtimov, Fernandes, Li, Rahmati, Xiao,
  Prakash, Kohno, and Song]{driving}
Kevin Eykholt, Ivan Evtimov, Earlence Fernandes, Bo~Li, Amir Rahmati, Chaowei
  Xiao, Atul Prakash, Tadayoshi Kohno, and Dawn Song.
\newblock Robust physical-world attacks on deep learning visual classification.
\newblock In \emph{CVPR}, pp.\  1625--1634, 2018.

\bibitem[Fischer \& Alemi(2020)Fischer and Alemi]{CEB}
Ian Fischer and Alexander~A. Alemi.
\newblock {CEB} improves model robustness.
\newblock \emph{Entropy}, 22\penalty0 (10):\penalty0 1081, 2020.

\bibitem[Fischer(2020)]{CEB2}
Ian~S. Fischer.
\newblock The conditional entropy bottleneck.
\newblock \emph{Entropy}, 22\penalty0 (9):\penalty0 999, 2020.

\bibitem[Freeman(1994)]{GenericVisualPerception}
William~T Freeman.
\newblock The generic viewpoint assumption in a framework for visual
  perception.
\newblock \emph{Nature}, 368\penalty0 (6471):\penalty0 542--545, 1994.

\bibitem[Goodfellow et~al.(2014)Goodfellow, Pouget{-}Abadie, Mirza, Xu,
  Warde{-}Farley, Ozair, Courville, and Bengio]{explaingingOfAdverGoodfellow2}
Ian~J. Goodfellow, Jean Pouget{-}Abadie, Mehdi Mirza, Bing Xu, David
  Warde{-}Farley, Sherjil Ozair, Aaron~C. Courville, and Yoshua Bengio.
\newblock Generative adversarial nets.
\newblock In Zoubin Ghahramani, Max Welling, Corinna Cortes, Neil~D. Lawrence,
  and Kilian~Q. Weinberger (eds.), \emph{NeurIPS}, pp.\  2672--2680, 2014.

\bibitem[Goodfellow et~al.(2015)Goodfellow, Shlens, and
  Szegedy]{explaingingOfAdverGoodfellow}
Ian~J. Goodfellow, Jonathon Shlens, and Christian Szegedy.
\newblock Explaining and harnessing adversarial examples.
\newblock In Yoshua Bengio and Yann LeCun (eds.), \emph{ICLR}, 2015.

\bibitem[Haralick et~al.(1973)Haralick, Shanmugam, and
  Dinstein]{imageClassification}
Robert~M. Haralick, Karthikeyan~S. Shanmugam, and Its'hak Dinstein.
\newblock Textural features for image classification.
\newblock \emph{{IEEE} Trans. Syst. Man Cybern.}, 3\penalty0 (6):\penalty0
  610--621, 1973.

\bibitem[He et~al.(2016)He, Zhang, Ren, and Sun]{ResNet}
Kaiming He, Xiangyu Zhang, Shaoqing Ren, and Jian Sun.
\newblock Deep residual learning for image recognition.
\newblock In \emph{CVPR}, pp.\  770--778, 2016.

\bibitem[Hinton et~al.(2015)Hinton, Vinyals, and Dean]{hinton2015distilling}
Geoffrey~E. Hinton, Oriol Vinyals, and Jeffrey Dean.
\newblock Distilling the knowledge in a neural network.
\newblock \emph{CoRR}, abs/1503.02531, 2015.

\bibitem[Kim et~al.(2022)Kim, Lee, and Ro]{DistillingRobustandNonRubust}
Junho Kim, Byung{-}Kwan Lee, and Yong~Man Ro.
\newblock Distilling robust and non-robust features in adversarial examples by
  information bottleneck.
\newblock \emph{CoRR}, abs/2204.02735, 2022.

\bibitem[Kingma \& Welling(2014)Kingma and Welling]{rePara}
Diederik~P. Kingma and Max Welling.
\newblock Auto-encoding variational bayes.
\newblock In Yoshua Bengio and Yann LeCun (eds.), \emph{ICLR}, 2014.

\bibitem[Korshunova et~al.(2021)Korshunova, Stutz, Alemi, Wiles, and
  Gowal]{closerlook}
Iryna Korshunova, David Stutz, Alexander~A. Alemi, Olivia Wiles, and Sven
  Gowal.
\newblock A closer look at the adversarial robustness of information bottleneck
  models.
\newblock \emph{CoRR}, abs/2107.05712, 2021.

\bibitem[Krizhevsky et~al.(2009)Krizhevsky, Hinton,
  et~al.]{krizhevsky2009learning}
Alex Krizhevsky, Geoffrey Hinton, et~al.
\newblock Learning multiple layers of features from tiny images.
\newblock 2009.

\bibitem[Krizhevsky et~al.(2017)Krizhevsky, Sutskever, and Hinton]{alexnet}
Alex Krizhevsky, Ilya Sutskever, and Geoffrey~E. Hinton.
\newblock Imagenet classification with deep convolutional neural networks.
\newblock \emph{Commun. {ACM}}, 60\penalty0 (6):\penalty0 84--90, 2017.
\newblock \doi{10.1145/3065386}.
\newblock URL \url{http://doi.acm.org/10.1145/3065386}.

\bibitem[Kuang et~al.(2020)Kuang, Li, Cui, Liu, Tao, Zhuang, and
  Wu]{StablePrediction}
Kun Kuang, Bo~Li, Peng Cui, Yue Liu, Jianrong Tao, Yueting Zhuang, and Fei Wu.
\newblock Stable prediction via leveraging seed variable.
\newblock \emph{CoRR}, abs/2006.05076, 2020.

\bibitem[Kusner et~al.(2017)Kusner, Loftus, Russell, and
  Silva]{counterfactualFairness}
Matt~J. Kusner, Joshua~R. Loftus, Chris Russell, and Ricardo Silva.
\newblock Counterfactual fairness.
\newblock \emph{CoRR}, abs/1703.06856, 2017.

\bibitem[LeCun et~al.(1998)LeCun, Bottou, Bengio, and Haffner]{MNISTref}
Yann LeCun, L{\'{e}}on Bottou, Yoshua Bengio, and Patrick Haffner.
\newblock Gradient-based learning applied to document recognition.
\newblock \emph{Proc. {IEEE}}, 86\penalty0 (11):\penalty0 2278--2324, 1998.

\bibitem[Lee et~al.(2017)Lee, Han, and Lee]{lee2017generative}
Hyeungill Lee, Sungyeob Han, and Jungwoo Lee.
\newblock Generative adversarial trainer: Defense to adversarial perturbations
  with {GAN}.
\newblock \emph{CoRR}, abs/1705.03387, 2017.

\bibitem[Liu et~al.(2020)Liu, Wang, Liu, Cao, Zhang, and Yu]{onlineShopping}
Aishan Liu, Jiakai Wang, Xianglong Liu, Bowen Cao, Chongzhi Zhang, and Hang Yu.
\newblock Bias-based universal adversarial patch attack for automatic
  check-out.
\newblock In \emph{ECCV}, pp.\  395--410, 2020.

\bibitem[Madry et~al.(2018)Madry, Makelov, Schmidt, Tsipras, and
  Vladu]{TowardsAdver}
Aleksander Madry, Aleksandar Makelov, Ludwig Schmidt, Dimitris Tsipras, and
  Adrian Vladu.
\newblock Towards deep learning models resistant to adversarial attacks.
\newblock In \emph{ICLR}, 2018.

\bibitem[Miyato et~al.(2017)Miyato, Dai, and Goodfellow]{miyato2016adversarial}
Takeru Miyato, Andrew~M. Dai, and Ian~J. Goodfellow.
\newblock Adversarial training methods for semi-supervised text classification.
\newblock In \emph{ICLR}, 2017.

\bibitem[Moosavi{-}Dezfooli et~al.(2017)Moosavi{-}Dezfooli, Fawzi, Fawzi, and
  Frossard]{UniversalAdverPertu}
Seyed{-}Mohsen Moosavi{-}Dezfooli, Alhussein Fawzi, Omar Fawzi, and Pascal
  Frossard.
\newblock Universal adversarial perturbations.
\newblock In \emph{CVPR}, pp.\  86--94, 2017.

\bibitem[Pan et~al.(2021)Pan, Niu, Zhang, and Zhang]{DisenIB}
Ziqi Pan, Li~Niu, Jianfu Zhang, and Liqing Zhang.
\newblock Disentangled information bottleneck.
\newblock In \emph{AAAI}, pp.\  9285--9293, 2021.

\bibitem[Parascandolo et~al.(2017)Parascandolo, Rojas{-}Carulla, Kilbertus, and
  Sch{\"{o}}lkopf]{learningCausalMechanisms}
Giambattista Parascandolo, Mateo Rojas{-}Carulla, Niki Kilbertus, and Bernhard
  Sch{\"{o}}lkopf.
\newblock Learning independent causal mechanisms.
\newblock \emph{CoRR}, abs/1712.00961, 2017.

\bibitem[Pearl(2009)]{CausalityPearl}
Judea Pearl.
\newblock \emph{Causality}.
\newblock Cambridge university press, 2009.

\bibitem[Pearl(2010)]{CausalInference}
Judea Pearl.
\newblock Causal inference.
\newblock In Isabelle Guyon, Dominik Janzing, and Bernhard Sch{\"{o}}lkopf
  (eds.), \emph{NIPS}, volume~6 of \emph{{JMLR} Proceedings}, pp.\  39--58,
  2010.

\bibitem[Peters et~al.(2016)Peters, B{\"u}hlmann, and
  Meinshausen]{CausalInferencebyUsingInvariantPrediction}
Jonas Peters, Peter B{\"u}hlmann, and Nicolai Meinshausen.
\newblock Causal inference by using invariant prediction: identification and
  confidence intervals.
\newblock \emph{Journal of the Royal Statistical Society: Series B (Statistical
  Methodology)}, 78\penalty0 (5):\penalty0 947--1012, 2016.

\bibitem[Peters et~al.(2017)Peters, Janzing, and
  Sch{\"o}lkopf]{ElementsofCausalInference}
Jonas Peters, Dominik Janzing, and Bernhard Sch{\"o}lkopf.
\newblock \emph{Elements of causal inference: foundations and learning
  algorithms}.
\newblock The MIT Press, 2017.

\bibitem[Pinot et~al.(2019)Pinot, Meunier, Araujo, Kashima, Yger,
  Gouy{-}Pailler, and Atif]{TheoreticalEvidenceAdver}
Rafael Pinot, Laurent Meunier, Alexandre Araujo, Hisashi Kashima, Florian Yger,
  C{\'{e}}dric Gouy{-}Pailler, and Jamal Atif.
\newblock Theoretical evidence for adversarial robustness through
  randomization.
\newblock In \emph{NeurIPS}, pp.\  11838--11848, 2019.

\bibitem[Redmon et~al.(2016)Redmon, Divvala, Girshick, and
  Farhadi]{objectDetection}
Joseph Redmon, Santosh~Kumar Divvala, Ross~B. Girshick, and Ali Farhadi.
\newblock You only look once: Unified, real-time object detection.
\newblock In \emph{CVPR}, pp.\  779--788, 2016.

\bibitem[Rifai et~al.(2011)Rifai, Vincent, Muller, Glorot, and
  Bengio]{rifai2011contractive}
Salah Rifai, Pascal Vincent, Xavier Muller, Xavier Glorot, and Yoshua Bengio.
\newblock Contractive auto-encoders: Explicit invariance during feature
  extraction.
\newblock In Lise Getoor and Tobias Scheffer (eds.), \emph{ICML}, pp.\
  833--840, 2011.

\bibitem[Sch{\"o}lkopf et~al.(2021)Sch{\"o}lkopf, Locatello, Bauer, Ke,
  Kalchbrenner, Goyal, and Bengio]{TowardCausalRepresentation}
Bernhard Sch{\"o}lkopf, Francesco Locatello, Stefan Bauer, Nan~Rosemary Ke, Nal
  Kalchbrenner, Anirudh Goyal, and Yoshua Bengio.
\newblock Toward causal representation learning.
\newblock \emph{Proceedings of the IEEE}, 109\penalty0 (5):\penalty0 612--634,
  2021.

\bibitem[Sharif et~al.(2016)Sharif, Bhagavatula, Bauer, and Reiter]{FaceRec}
Mahmood Sharif, Sruti Bhagavatula, Lujo Bauer, and Michael~K. Reiter.
\newblock Accessorize to a crime: Real and stealthy attacks on state-of-the-art
  face recognition.
\newblock In Edgar~R. Weippl, Stefan Katzenbeisser, Christopher Kruegel,
  Andrew~C. Myers, and Shai Halevi (eds.), \emph{SIGSAC}, pp.\  1528--1540,
  2016.

\bibitem[Shin \& Song(2017)Shin and Song]{shin2017jpeg}
Richard Shin and Dawn Song.
\newblock Jpeg-resistant adversarial images.
\newblock In \emph{NIPS}, volume~1, 2017.

\bibitem[Shwartz{-}Ziv \& Tishby(2017)Shwartz{-}Ziv and
  Tishby]{OpeningBBTishby}
Ravid Shwartz{-}Ziv and Naftali Tishby.
\newblock Opening the black box of deep neural networks via information.
\newblock \emph{CoRR}, abs/1703.00810, 2017.

\bibitem[Sinha et~al.(2021)Sinha, Bharadhwaj, Goyal, Larochelle, Garg, and
  Shkurti]{DIBS}
Samarth Sinha, Homanga Bharadhwaj, Anirudh Goyal, Hugo Larochelle, Animesh
  Garg, and Florian Shkurti.
\newblock {DIBS:} diversity inducing information bottleneck in model ensembles.
\newblock In \emph{AAAI}, pp.\  9666--9674, 2021.

\bibitem[Szegedy et~al.(2014{\natexlab{a}})Szegedy, Zaremba, Sutskever, Bruna,
  Erhan, Goodfellow, and Fergus]{existOfAdverGoodfellow}
Christian Szegedy, Wojciech Zaremba, Ilya Sutskever, Joan Bruna, Dumitru Erhan,
  Ian~J. Goodfellow, and Rob Fergus.
\newblock Intriguing properties of neural networks.
\newblock In Yoshua Bengio and Yann LeCun (eds.), \emph{ICLR},
  2014{\natexlab{a}}.

\bibitem[Szegedy et~al.(2014{\natexlab{b}})Szegedy, Zaremba, Sutskever, Bruna,
  Erhan, Goodfellow, and Fergus]{existOfAdverSzegedy}
Christian Szegedy, Wojciech Zaremba, Ilya Sutskever, Joan Bruna, Dumitru Erhan,
  Ian~J. Goodfellow, and Rob Fergus.
\newblock Intriguing properties of neural networks.
\newblock In Yoshua Bengio and Yann LeCun (eds.), \emph{ICLR},
  2014{\natexlab{b}}.

\bibitem[Tang et~al.(2021)Tang, Tao, and Zhang]{CiiV}
Kaihua Tang, Mingyuan Tao, and Hanwang Zhang.
\newblock Adversarial visual robustness by causal intervention.
\newblock \emph{CoRR}, abs/2106.09534, 2021.

\bibitem[Tishby \& Zaslavsky(2015)Tishby and Zaslavsky]{DLTishby}
Naftali Tishby and Noga Zaslavsky.
\newblock Deep learning and the information bottleneck principle.
\newblock In \emph{ITW}, pp.\  1--5, 2015.

\bibitem[Tishby et~al.(2000)Tishby, Pereira, and Bialek]{IBTishby}
Naftali Tishby, Fernando C.~N. Pereira, and William Bialek.
\newblock The information bottleneck method.
\newblock \emph{CoRR}, physics/0004057, 2000.

\bibitem[Tram{\`{e}}r et~al.(2018)Tram{\`{e}}r, Kurakin, Papernot, Goodfellow,
  Boneh, and McDaniel]{EnsembleAdversarialTraining}
Florian Tram{\`{e}}r, Alexey Kurakin, Nicolas Papernot, Ian~J. Goodfellow, Dan
  Boneh, and Patrick~D. McDaniel.
\newblock Ensemble adversarial training: Attacks and defenses.
\newblock In \emph{ICLR}, 2018.

\bibitem[Voloshynovskiy et~al.(2019)Voloshynovskiy, Kondah, Rezaeifar, Taran,
  Holotyak, and Rezende]{variationalGlass}
Slava Voloshynovskiy, Mouad Kondah, Shideh Rezaeifar, Olga Taran, Taras
  Holotyak, and Danilo~Jimenez Rezende.
\newblock Information bottleneck through variational glasses.
\newblock \emph{CoRR}, abs/1912.00830, 2019.

\bibitem[Wang et~al.(2016)Wang, Guo, Zhang, II, Xing, Giles, and
  Liu]{RandomAdver}
Qinglong Wang, Wenbo Guo, Kaixuan Zhang, Alexander G.~Ororbia II, Xinyu Xing,
  C.~Lee Giles, and Xue Liu.
\newblock Learning adversary-resistant deep neural networks.
\newblock \emph{CoRR}, abs/1612.01401, 2016.

\bibitem[Warde{-}Farley \& Bengio(2017)Warde{-}Farley and
  Bengio]{ImproveWithDenoising}
David Warde{-}Farley and Yoshua Bengio.
\newblock Improving generative adversarial networks with denoising feature
  matching.
\newblock In \emph{ICLR}, 2017.

\bibitem[Xiao et~al.(2017)Xiao, Rasul, and Vollgraf]{fashionMNISTref}
Han Xiao, Kashif Rasul, and Roland Vollgraf.
\newblock Fashion-mnist: a novel image dataset for benchmarking machine
  learning algorithms.
\newblock \emph{CoRR}, abs/1708.07747, 2017.

\bibitem[Xie et~al.(2017)Xie, Wang, Zhang, Zhou, Xie, and
  Yuille]{AdverSemanticSeg}
Cihang Xie, Jianyu Wang, Zhishuai Zhang, Yuyin Zhou, Lingxi Xie, and Alan
  Yuille.
\newblock Adversarial examples for semantic segmentation and object detection.
\newblock In \emph{ICCV}, pp.\  1369--1378, 2017.

\bibitem[Xie et~al.(2019)Xie, Wu, van~der Maaten, Yuille, and
  He]{FeatureDenoising}
Cihang Xie, Yuxin Wu, Laurens van~der Maaten, Alan~L. Yuille, and Kaiming He.
\newblock Feature denoising for improving adversarial robustness.
\newblock In \emph{CVPR}, pp.\  501--509, 2019.

\bibitem[Xu et~al.(2020)Xu, Ma, Liu, Deb, Liu, Tang, and Jain]{reviewOfAdverXu}
Han Xu, Yao Ma, Haochen Liu, Debayan Deb, Hui Liu, Jiliang Tang, and Anil~K.
  Jain.
\newblock Adversarial attacks and defenses in images, graphs and text: {A}
  review.
\newblock \emph{Int. J. Autom. Comput.}, 17\penalty0 (2):\penalty0 151--178,
  2020.

\bibitem[Zagoruyko \& Komodakis(2016)Zagoruyko and Komodakis]{WideResNet}
Sergey Zagoruyko and Nikos Komodakis.
\newblock Wide residual networks.
\newblock In \emph{BMVC}, 2016.

\bibitem[Zhang et~al.(2020)Zhang, Zhang, and Li]{CausalViewRobust}
Cheng Zhang, Kun Zhang, and Yingzhen Li.
\newblock A causal view on robustness of neural networks.
\newblock In Hugo Larochelle, Marc'Aurelio Ranzato, Raia Hadsell,
  Maria{-}Florina Balcan, and Hsuan{-}Tien Lin (eds.), \emph{NeurIPS}, 2020.

\bibitem[Zhang et~al.(2018)Zhang, Ciss{\'{e}}, Dauphin, and Lopez{-}Paz]{mixup}
Hongyi Zhang, Moustapha Ciss{\'{e}}, Yann~N. Dauphin, and David Lopez{-}Paz.
\newblock mixup: Beyond empirical risk minimization.
\newblock In \emph{ICLR}, 2018.

\bibitem[Zhang et~al.(2021)Zhang, Gong, Liu, Niu, Tian, Han, Sch{\"{o}}lkopf,
  and Zhang]{AdverThrougnLen}
Yonggang Zhang, Mingming Gong, Tongliang Liu, Gang Niu, Xinmei Tian, Bo~Han,
  Bernhard Sch{\"{o}}lkopf, and Kun Zhang.
\newblock Adversarial robustness through the lens of causality.
\newblock \emph{CoRR}, abs/2106.06196, 2021.

\bibitem[Zheng et~al.(2016)Zheng, Song, Leung, and
  Goodfellow]{zheng2016improving}
Stephan Zheng, Yang Song, Thomas Leung, and Ian~J. Goodfellow.
\newblock Improving the robustness of deep neural networks via stability
  training.
\newblock In \emph{CVPR}, pp.\  4480--4488, 2016.

\end{thebibliography}
\bibliographystyle{iclr2023_conference}

\appendix
\section{Appendix}
\subsection{Information Bottleneck}
In this section, we specifically describe how VIB uses variational inference and reparameterization trick to optimize information bottleneck (IB) in DNNs.

\begin{figure}[t]
    \centering
    \includegraphics[width=0.3\linewidth]{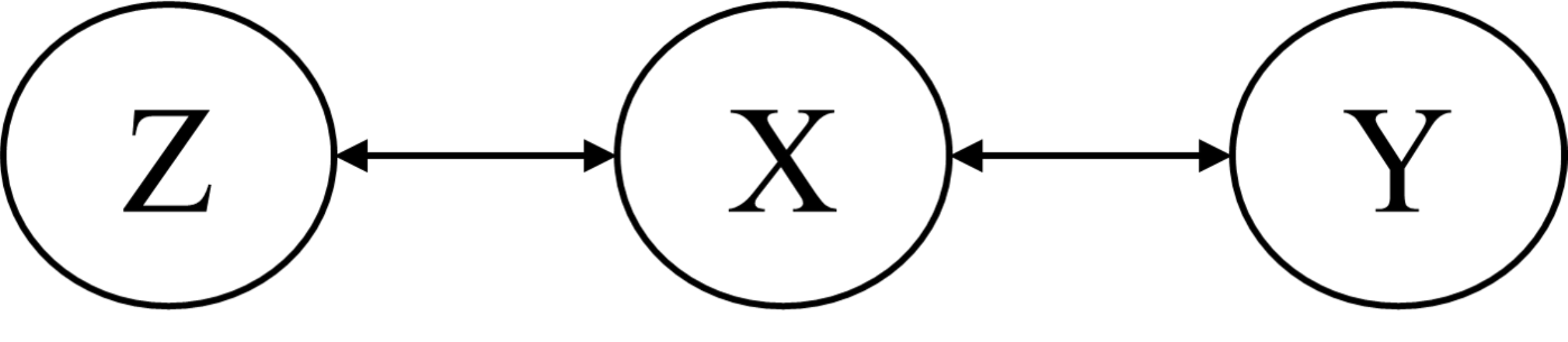}
    \caption{Markov Chain.}
    \label{Markov}
\end{figure}

Gievn a Markov Chain shown in Figure~\ref{Markov}, where $X$ represents the source random variable, $Y$ represents the target random variable and $Z$ represents the compression code. The purpose of IB is to maximize the objective loss function as Eq.\ref{AppendixOriginIBLoss}:
\begin{equation}\label{AppendixOriginIBLoss}
\mathcal{L}_{IB}=-I\left( Z;Y \right) +\alpha I\left( Z;X \right). 
\end{equation}
$I\left( Z;Y \right)$ can be written as the form of integral defined in Eq.\ref{IZY}:
\begin{equation}\label{IZY}
I(Z, Y)=\int d y d z p(y, z) \log \frac{p(y, z)}{p(y) p(z)}=\int d y d z p(y, z) \log \frac{p(y \mid z)}{p(y)},
\end{equation}
where $p(y \mid z)$ is defined by Markov Chain $Y \leftrightarrow X \leftrightarrow Z$ as Eq.\ref{pyz}:
\begin{equation}\label{pyz}
p(y \mid z)=\int d x p(x, y \mid z)=\int d x p(y \mid x) p(x \mid z)=\int d x \frac{p(y \mid x) p(z \mid x) p(x)}{p(z)}.
\end{equation}
In fact, $p(y \mid z)$ is intractable, so $q(y \mid z)$ is needed to approximate $p(y \mid z)$. Using the fact that the Kullback Leibler divergence is always positive, Eq.\ref{bala} can be induced:
\begin{equation}\label{bala}
\mathrm{KL}[p(Y \mid Z), q(Y \mid Z)] \geq 0 \Longrightarrow \int d y p(y \mid z) \log p(y \mid z) \geq \int d y p(y \mid z) \log q(y \mid z),
\end{equation}
and hence, $I(Z,Y)$ can be induced as Eq.\ref{nnewIZY}:
\begin{equation}\label{nnewIZY}
    \begin{aligned}
        I(Z,Y)&\ge \int{d}ydzp(y,z)\log \frac{q(y\mid z)}{p(y)}\\
              &=\int{d}ydzp(y,z)\log q(y\mid z)-\int{d}yp(y)\log p(y)\\
              &=\int{d}ydzp(y,z)\log q(y\mid z)+H(Y). 
    \end{aligned}
\end{equation}
Leveraging the Markov assumption, $p(y, z)$ can be written as Eq.\ref{newpyz}:
\begin{equation}\label{newpyz}
p(y,z)=\int{dxp\left( x,y,z \right) =\int{dxp\left( x \right) p\left( y|x \right) p\left( z|x \right)}},
\end{equation}
which gives us a new lower bound on the first term of our objective:
\begin{equation}
I(Z, Y) \geq \int d x d y d z p(x) p(y \mid x) p(z \mid x) \log q(y \mid z).
\end{equation}
Then, the expression $I\left( Z;X \right)$ can be written as Eq.\ref{IZXChange}:
\begin{equation}\label{IZXChange}
I(Z, X)=\int d z d x p(x, z) \log \frac{p(z \mid x)}{p(z)}=\int d z d x p(x, z) \log p(z \mid x)-\int d z p(z) \log p(z).
\end{equation}
In general, while it is fully defined, computing the marginal distribution of $p\left( z \right) =\int{dxp\left( z|x \right) p\left( x \right)}$ might be difficult. Let $r(z)$ be a variational approximation to this marginal, the upper bound defined in Eq.\ref{IZXDeform}:
\begin{equation}\label{IZXDeform}
I(Z, X) \leq \int d x d z p(x) p(z \mid x) \log \frac{p(z \mid x)}{r(z)}.
\end{equation}
Combining both of these bounds, Eq.\ref{gaga} can be induced:
\begin{equation}\label{gaga}
\begin{aligned}
I(Z, Y)-\alpha I(Z, X) \geq & \int d x d y d z p(x) p(y \mid x) p(z \mid x) \log q(y \mid z) \\
&-\alpha \int d x d z p(x) p(z \mid x) \log \frac{p(z \mid x)}{r(z)}=L.
\end{aligned}
\end{equation}
$p\left( x,y \right)$ can be approximated using the empirical data distribution $p(x, y)=\frac{1}{N} \sum_{n=1}^{N} \delta_{x_{n}}(x) \delta_{y_{n}}(y)$,  so that Eq.\ref{firstL} can be induced:
\begin{equation}\label{firstL}
L \approx \frac{1}{N} \sum_{n=1}^{N}\left[\int d z p\left(z \mid x_{n}\right) \log q\left(y_{n} \mid z\right)-\alpha p\left(z \mid x_{n}\right) \log \frac{p\left(z \mid x_{n}\right)}{r(z)}\right].
\end{equation}
Suppose there exists an encoder of the form $p(z \mid x)=\mathcal{N}\left(z \mid f_{e}^{\mu}(x)\right.$, where $f_{e}$ is the VIB model, then, the reparameterization trick can be utilized to do the transformation $p(z \mid x) d z=p(\epsilon) d \epsilon$, where $z=f(x, \epsilon)$ is a deterministic function of $x$, and the Gaussian random variable $\epsilon$ and hence the total loss function $L_{IB}$ can be induced as Eq.\ref{totalIBlossAppendix}:
\begin{equation}\label{totalIBlossAppendix}
L_{I B}=\frac{1}{N} \sum_{n=1}^{N} \mathbb{E}_{\epsilon \sim p(\epsilon)}\left[-\log q\left(y_{n} \mid f\left(x_{n}, \epsilon\right)\right)\right]+\alpha \operatorname{KL}\left[p\left(Z \mid x_{n}\right), r(Z)\right],
\end{equation}
and the former conditional entropy can be approximated using cross entropy.
\subsection{Visualization}
\begin{figure}[!t]
	\centering
	\begin{minipage}{0.49\linewidth}
		\centering
		\includegraphics[width=1\linewidth]{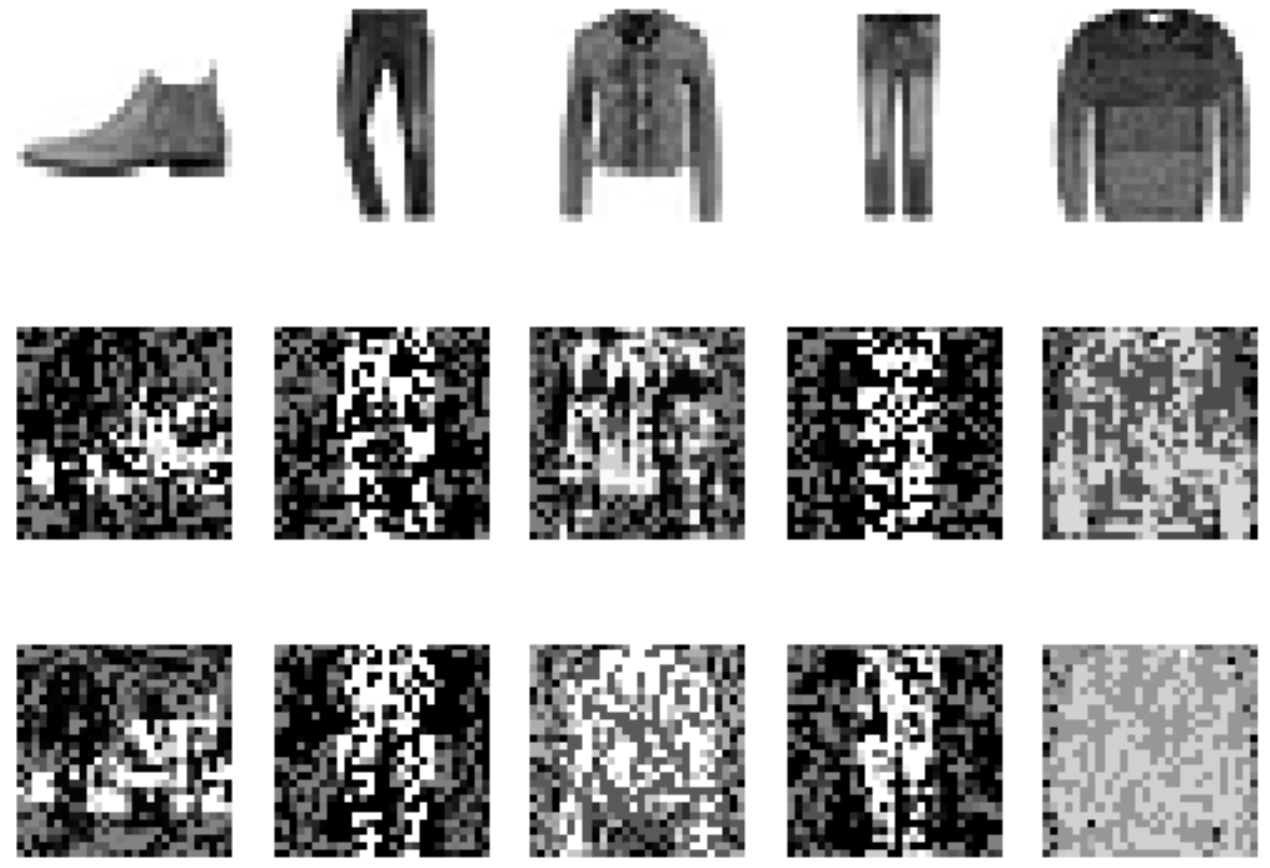}
	\end{minipage}
	%\qquad
	\begin{minipage}{0.49\linewidth}
		\centering
		\includegraphics[width=1\linewidth]{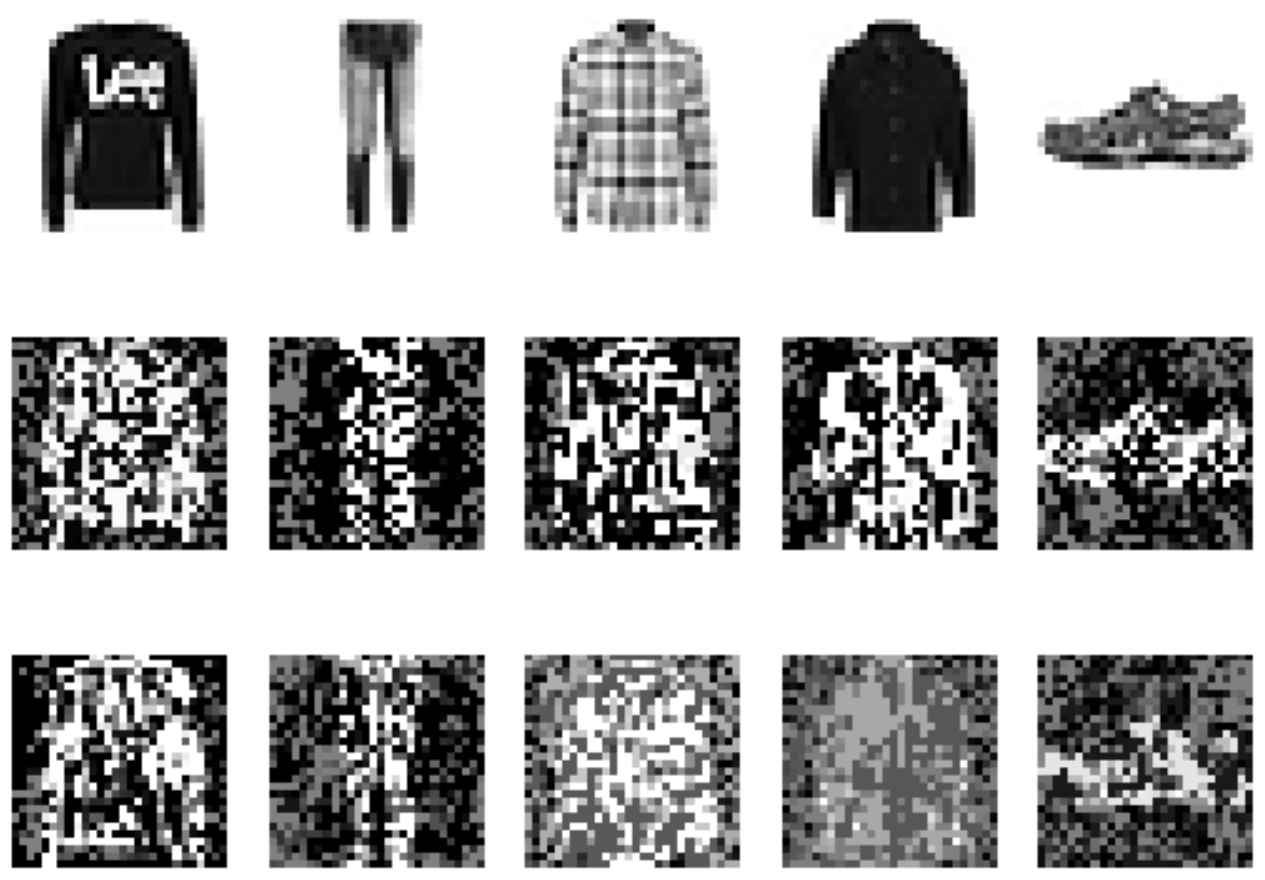}
	\end{minipage}
\caption{Visualization Examples 1 on the dataset of FashionMNIST~\cite{fashionMNISTref}.}
\label{appendixfig1}
\end{figure}

\begin{figure}[!t]
	\centering
	\begin{minipage}{0.49\linewidth}
		\centering
		\includegraphics[width=1\linewidth]{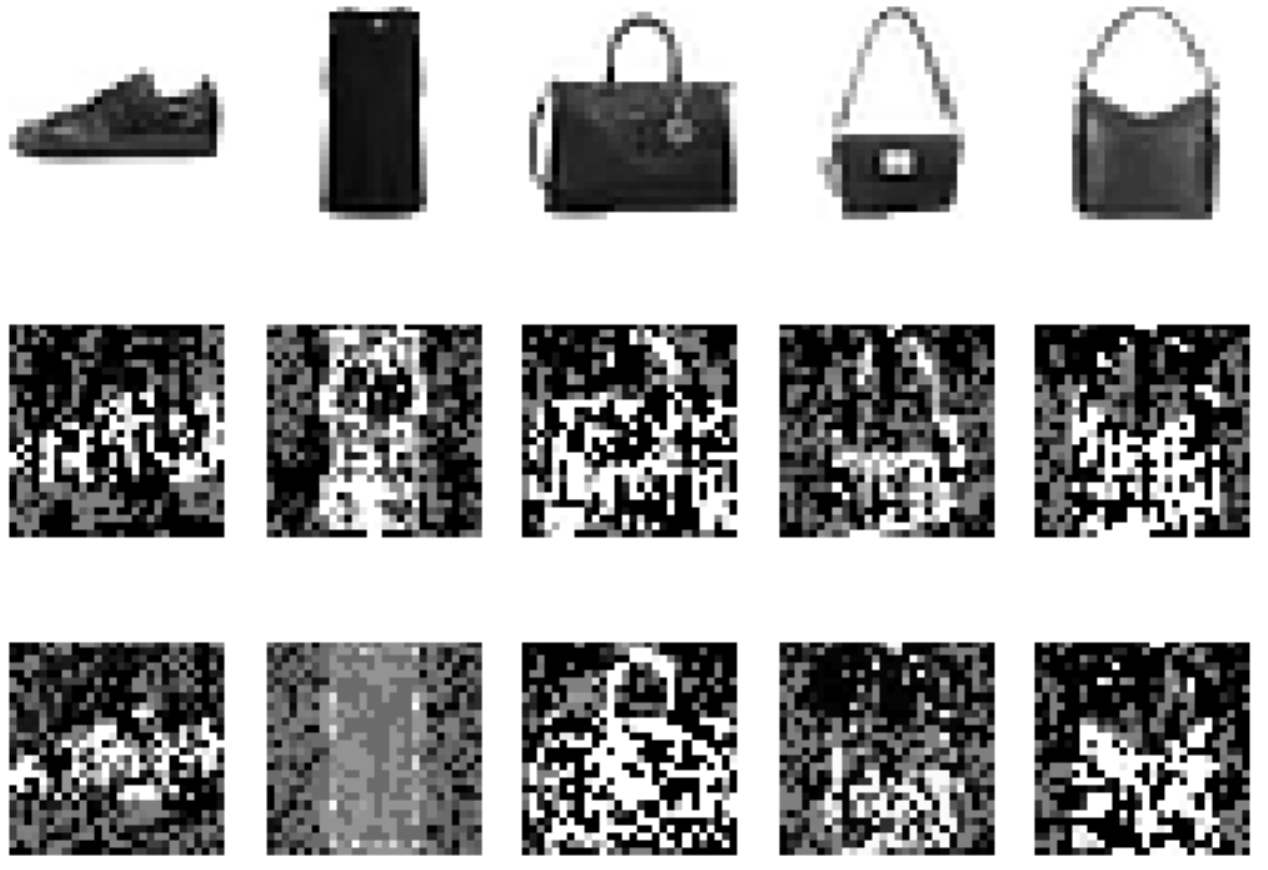}
	\end{minipage}
	%\qquad
	\begin{minipage}{0.49\linewidth}
		\centering
		\includegraphics[width=1\linewidth]{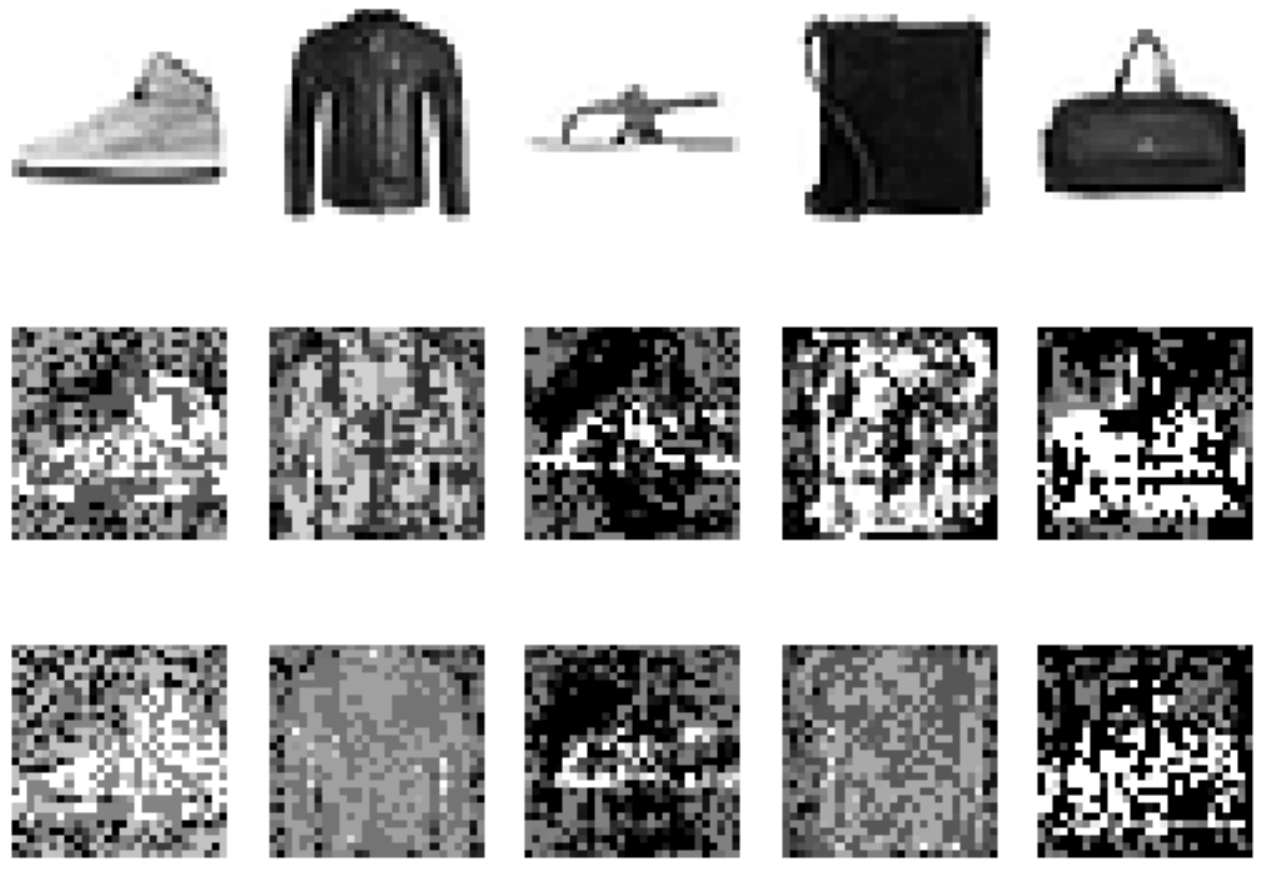}
	\end{minipage}
\caption{Visualization Examples 2 on the dataset of FashionMNIST~\cite{fashionMNISTref}.}
\label{appendixfig2}
\end{figure}

To verify that the proposed CausalIB has a better performance than VIB in feature extraction, we make the visualization of adversarial examples as shown in Figure~\ref{appendixfig1} and Figure~\ref{appendixfig2}.

In Figure~\ref{appendixfig1} and Figure~\ref{appendixfig2}, the visualization of clean examples are shown in the first row. The visualization of adversarial examples attacking the CausalIB model and the visualization of adversarial examples attacking the VIB model are shown in the second and third rows. The adversarial examples attacking the VIB model are relatively blurry, while the adversarial examples attacking the CausalIB model have much clearer structural details. This means that the causal inference method could lead the model to learn more structural information. As a result, the adversary must add larger perturbations with the erasion of the structural patterns to fool the CausalIB model. 

\subsection{More Detailed Ablation Experiments}
In this section, more detailed ablation experiment results are provided to show the impact of different choices of the two hyperparameters in our method. 
\begin{table}[!t]
\centering
\scriptsize
\caption{More detailed ablation experiments.}
\label{ablationtable}
\begin{tabular}{llll}
\hline
\textbf{Method}                                       & \textbf{clean}              & \textbf{FGSM}               & \textbf{PGD-20}             \\ \hline
$\alpha$=0.01/ $\beta$=0.01                           & 95.32\thinspace$\pm$\thinspace0.98 & 73.50\thinspace$\pm$\thinspace0.55 & 34.41\thinspace$\pm$\thinspace0.85 \\
$\alpha$=0.01/ $\beta$=0.2                            & 96.24\thinspace$\pm$\thinspace1.27 & 71.27\thinspace$\pm$\thinspace0.74 & 33.90\thinspace$\pm$\thinspace1.36 \\
$\alpha$=0.01/ $\beta$=1.0  & 90.35\thinspace$\pm$\thinspace0.91 & 60.23\thinspace$\pm$\thinspace1.52 & 28.73\thinspace$\pm$\thinspace0.89 \\
$\alpha$=0.05/ $\beta$=0.01 & 97.46\thinspace$\pm$\thinspace0.47 & 78.40\thinspace$\pm$\thinspace0.30 & 50.42\thinspace$\pm$\thinspace0.28 \\
$\alpha$=0.05/ $\beta$=0.2  & 96.39\thinspace$\pm$\thinspace0.76 & 76.52\thinspace$\pm$\thinspace0.77 & 49.69\thinspace$\pm$\thinspace0.61 \\
$\alpha$=0.05/ $\beta$=0.5  & 93.56\thinspace$\pm$\thinspace0.58 & 69.09\thinspace$\pm$\thinspace1.39 & 29.30\thinspace$\pm$\thinspace1.25 \\
$\alpha$=0/ $\beta$=0.05    & 94.62\thinspace$\pm$\thinspace0.56 & 42.99\thinspace$\pm$\thinspace0.72 & 16.88\thinspace$\pm$\thinspace0.88 \\
$\alpha$=0/ $\beta$=0.2     & 95.28\thinspace$\pm$\thinspace0.64 & 38.49\thinspace$\pm$\thinspace0.59 & 15.33\thinspace$\pm$\thinspace0.96 \\
$\alpha$=0/ $\beta$=1.0     & 92.52\thinspace$\pm$\thinspace0.45 & 26.20\thinspace$\pm$\thinspace0.87 & 12.16\thinspace$\pm$\thinspace1.68
\end{tabular}
\end{table}
Many adversarial defense methods have a trade-off between clean accuracy and adversarial robustness. When the weight of the regularization term is too large, the learning of the classifier could be hurt. As shown in Table \ref{ablationtable}, it can be seen that a too-large $\beta$ or a too-small $\beta$ will lead to a decrease in the defense performance of the model against adversarial examples. 

Currently, the settings of the hyperparameters are empirical, which is a shortcoming of the proposed CausalIB. In the future, a more reasonable hyperparameter optimization method would be studied.

\subsection{Experiments on CIFAR-100}
\begin{table}[!t]
\centering
\scriptsize
\caption{Experiment results on CIFAR-100.}
\begin{tabular}{ccccc}
\hline
           & clean          & FGSM           & PGD-20         & CW-20          \\
\hline
ResNet w/o defense~\cite{ResNet} & \textbf{70.10} & 3.18           & 0.0            & 0.0            \\
VIB~\cite{VIB}        & 56.48          & 21.48          & 12.77          & 10.04          \\
CiiV~\cite{CiiV}       & 52.15          & \textbf{32.51} & 23.19          & 22.05          \\
causalIB   & 52.58          & 32.06          & \textbf{24.85} & \textbf{22.52}
\end{tabular}
\label{CIFAR-100TABLE}
\end{table}
\textcolor{black}{In this section, results on the CIFAR-100 dataset are provided. The utilized model is ResNet34~\cite{ResNet}. VIB~\cite{VIB}, CiiV~\cite{CiiV}, and CausalIB are implemented on the ResNet model. The default ResNet model has no defense (no additional regularization method). As can be seen in Table \ref{CIFAR-100TABLE}, causalIB achieves the considerable adversarial robustness, but the classification accuracy on the clean data has a large drop, which is in line with the general trend of adversarial defense methods.}

\subsection{Experiment of Adaptive C\&W Attack}
\begin{table}[!t]
\centering
\scriptsize
\caption{The adversarial robustness under the C\&W attacks on different datasets.}
\begin{tabular}{ccccc}
\hline
           & MNIST          & FashionMNIST   & CIFAR-10       & CIFAR-100      \\
\hline
no defense & 88.92          & 52.65          & 1.28           & 0              \\
VIB~\cite{VIB}        & 94.45          & 86.05          & 18.04          & 10.45          \\
CiiV~\cite{CiiV}       & 96.08          & 90.15          & 32.59          & \textbf{24.86} \\
causalIB   & \textbf{96.55} & \textbf{90.77} & \textbf{35.44} & 23.93         
\end{tabular}
\label{CWtable}
\end{table}
\textcolor{black}{In this section, experiment results of the performances under the adaptive C\&W attacks~\cite{CW} are provided. The feature extraction model used on MNIST and FashionMNIST is a MLP, AlexNet~\cite{alexnet} is used on the CIFAR-10 dataset, and ResNet34~\cite{ResNet} is used on the CIFAR-100 dataset. The iteration step of C\&W attack is 20. As can be seen in Table \ref{CWtable}, the causalIB method still achieves the considerable adversarial robustness under the C\&W attack, which is consistent with the test results under the other attacks.}

\subsection{Comparison Experiments of AlexNet, ResNet, and WideResNet}
\begin{table}[!t]
\centering
\scriptsize
\caption{Comparison Experiments of AlexNet, ResNet and WideResNet on clean data and FGSM.}
\begin{tabular}{ccccccc}
\hline
        & \multicolumn{3}{c}{\textbf{clean}} & \multicolumn{3}{c}{\textbf{FGSM}} \\
\hline
        & AlexNet       & ResNet       & WideResNet            & AlexNet     & ResNet       & WideResNet            \\
\hline
No defense & 88.06        & 91.16        & \textbf{92.28}      & 30.48        & 39.12        & \textbf{39.85}     \\
VIB~\cite{VIB}      & 91.93        & \textbf{92.65}        & 92.45      & 41.89        & 44.2        & \textbf{46.79}      \\
CiiV~\cite{CiiV}     & 89.50        & 91.53        & \textbf{92.05}       & 52.95        & 56.18        & \textbf{58.25}     \\
causalIB & 91.76        & \textbf{92.91}        & 92.66      & 54.11        & 57.52        & \textbf{58.75} 
\end{tabular}
\label{tableCNN_a}
\end{table}
\begin{table}[!t]
\centering
\scriptsize
\caption{Comparison Experiments of AlexNet, ResNet, and WideResNet on PGD-20 and AutoAttack-$L_{inf}$.}
\begin{tabular}{ccccccc}
\hline
         & \multicolumn{3}{c}{\textbf{PGD-20}} & \multicolumn{3}{c}{\textbf{AA-$L_{inf}$}} \\
\hline
         & AlexNet       & ResNet       & WideResNet              & AlexNet     & ResNet       & WideResNet            \\
\hline
No defense & 0.12          & 0.75          & \textbf{2.59}       & 0           & 0           & 0        \\
VIB      & 22.81         & \textbf{27.47}         & 27.38      & 18.60        & 25.60        & \textbf{26.58}     \\
CiiV     & 33.62         & 42.30         & \textbf{43.66}       & 26.33       & 33.14       & \textbf{34.20}    \\
causalIB & 35.89         & \textbf{43.11}      & 43.05      & 28.7      & 34.32        & \textbf{34.66} 
\end{tabular}
\label{tableCNN_b}
\end{table}
\textcolor{black}{In this section, the experiment result of adversarial robustness influenced by different CNN structures on the CIFAR-10 dataset would be illustrated. Table \ref{tableCNN_a} and Table \ref{tableCNN_b} shows the comparative test results of the CNN models including AlexNet~\cite{alexnet}, ResNet34~\cite{ResNet}, and WideResNet-34-10~\cite{WideResNet}. It can be seen that our proposed method could achieve even better adversarial robustness in the residual structures~\cite{ResNet, WideResNet}, although the CausalIB has promoted adversarial robustness on the AlexNet model~\cite{alexnet}.}

\end{document}